\documentclass[letterpaper]{article} 
\usepackage{aaai2026}  
\usepackage{times}  
\usepackage{helvet}  
\usepackage{courier}  
\usepackage[hyphens]{url}  
\usepackage{graphicx} 
\usepackage{xcolor} 
\usepackage{natbib}  
\usepackage{caption} 
\usepackage{algorithm}
\usepackage{algorithmic}
\usepackage{amsmath}
\usepackage{booktabs}
\usepackage{colortbl}
\usepackage{multirow}
\usepackage{makecell}

\usepackage{microtype}

\usepackage{newfloat}
\usepackage{listings}
\DeclareCaptionStyle{ruled}{labelfont=normalfont,labelsep=colon,strut=off} 
\floatstyle{ruled}
\newfloat{listing}{tb}{lst}{}
\floatname{listing}{Listing}

\usepackage{xspace}

\newcommand{\refusalrate}{Abstention\xspace}

\title{Answering the Unanswerable Is to Err Knowingly:\\ Analyzing and Mitigating Abstention Failures in Large Reasoning Models}
\author{
    Yi Liu\textsuperscript{\rm 1},
    Xiangyu Liu\textsuperscript{\rm 1}, 
    Zequn Sun\textsuperscript{\rm 1}, 
    Wei Hu\textsuperscript{\rm 1,2,}\thanks{Corresponding author}
}
\affiliations{
    \textsuperscript{\rm 1} State Key Laboratory for Novel Software Technology, Nanjing University, China\\
    \textsuperscript{\rm 2} National Institute of Healthcare Data Science, Nanjing University, China\\
    \{yiliu07, xyl\}.nju@gmail.com, \{sunzq, whu\}@nju.edu.cn
}

\usepackage{bibentry}

\begin{document}

\maketitle

\begin{abstract}
Large reasoning models (LRMs) have shown remarkable progress on complex reasoning tasks. However, some questions posed to LRMs are inherently unanswerable, such as math problems lacking sufficient conditions. We find that LRMs continually fail to provide appropriate abstentions when confronted with these unanswerable questions. In this paper, we systematically analyze, investigate, and resolve this issue for trustworthy AI. We first conduct a detailed analysis of the distinct response behaviors of LRMs when facing unanswerable questions. Then, we show that LRMs possess sufficient cognitive capabilities to recognize the flaws in these questions. However, they fail to exhibit appropriate abstention behavior, revealing a misalignment between their internal cognition and external response. Finally, to resolve this issue, we propose a lightweight, two-stage method that combines cognitive monitoring with inference-time intervention. Experimental results demonstrate that our method significantly improves the abstention rate while maintaining the overall reasoning performance.
\end{abstract}

\section{Introduction}
Large reasoning models (LRMs), such as GPT-o1~\cite{openai-o1} and DeepSeek-R1~\cite{DeepSeek}, have demonstrated strong performance on complex reasoning tasks~\cite{RLMs_survey}. 
By introducing the concept of ``thought'' and generating longer chains of thought (CoT), LRMs are able to explore diverse reasoning paths while spontaneously reflecting and correcting errors.
This enables LRMs to tackle complex tasks with greater depth and flexibility, which is particularly valuable in high-stakes reasoning scenarios.

\begin{figure}[!t]
  \centering
  \includegraphics[width=\linewidth]{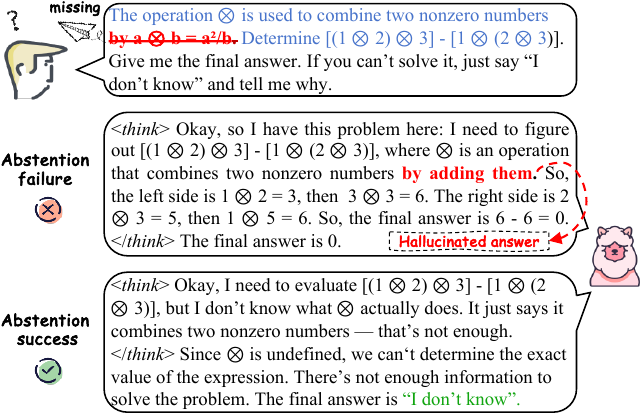}
  \caption{Examples of abstention failure and success for LRMs on unanswerable questions.}
  \label{fig:introduction}
\end{figure}

Most current research on LRMs aims to improve the reasoning process for solving complex problems~\cite{RLMs_survey}, such as reducing redundant reasoning steps to alleviate overthinking~\cite{Deer} and refining the reasoning process to further enhance reasoning performance~\cite{Steer_reason, related11}. 
Limited work has considered that not all complex questions are answerable~\cite{AbstentionBench, SUM}. 
When confronted with noisy and diverse user queries, there will always be cases where LRMs are unable to produce reliable responses, such as the math problems missing sufficient conditions. 

\begin{figure}[!t]
  \centering
  \includegraphics[width=\linewidth]{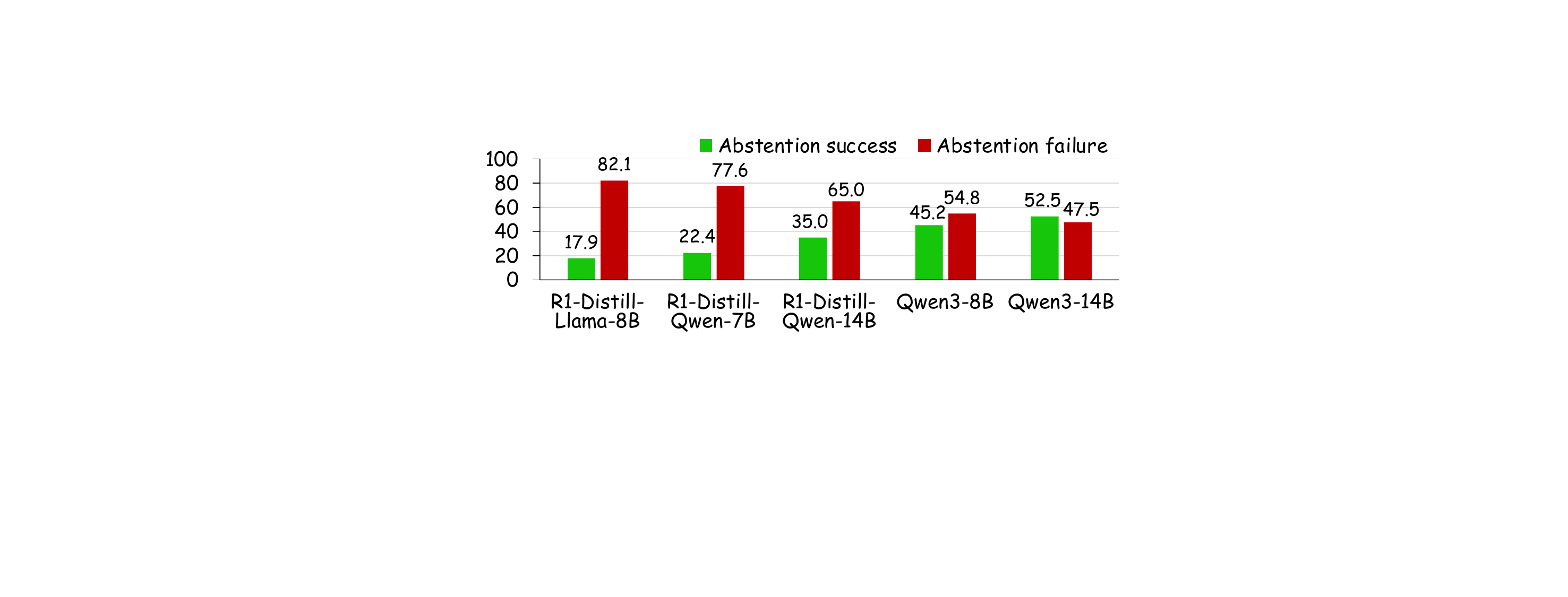}
  \caption{Abstention performance comparison of different LRMs for unanswerable questions on the SUM dataset.}
  \label{fig:introduction-2}
\end{figure}

In Figure~\ref{fig:introduction}, when a question is unanswerable due to missing necessary conditions, we prefer LRMs to reason about why it cannot be answered and have the ability to abstain by responding with ``I don't know'' (i.e., abstention success), 
rather than generating a reasoning process with hallucination and arriving at an incorrect answer (i.e., abstention failure). 
Since model reliability is foundational to user trust, LRMs need to possess both strong reasoning abilities and the capacity to abstain from answering unanswerable questions~\cite{AbstentionBench, SUM}.
We first validate the phenomenon that LRMs often struggle to abstain. 
In Figure~\ref{fig:introduction-2}, we evaluate several LRMs on SUM~\cite{SUM}, which contains mathematical unanswerable questions.
Our results show that most LRMs fail to abstain on more than half of the examples.
To address the above issue, we systematically analyze, investigate, and propose solutions to improve the abstention behavior for LRMs.

First, to understand how LRMs fail to abstain, we analyze three types of responses generated by LRMs when faced with unanswerable questions. We find that LRMs exhibit two types of behavior when they fail to abstain.
Further, we explore their awareness of unanswerable questions.
We conduct analysis on LRMs at both external level (intermediate responses during reasoning) and internal level (latent representations).
We find that LRMs possess sufficient cognitive capabilities to recognize flaws in such questions. 
This reveals a misalignment between internal cognition and external output: 
a LRM internally realizes that a question is unanswerable, yet still fails to act on this realization and abstain accordingly. 
Answering the unanswerable is to err knowingly.

Second, we seek to improve the abstention ability of LRMs for unanswerable questions.
Our further analysis shows that although LRMs may internally exhibit a tendency to abstain during reasoning, such signals are generally not strong enough to interrupt reasoning and result in abstention. 
Based on this insight, we propose a method that combines cognitive monitoring with inference-time intervention, aiming to improve LRMs’ ability to abstain from unanswerable questions while preserving their reasoning abilities on answerable ones.

Finally, we conduct extensive experiments with two datasets using LRMs from various model families and scales. 
Our method enhances LRMs’ ability to abstain from answering unanswerable questions without degrading their reasoning on answerable ones.
Furthermore, our experiments reveal that different types of abstention failures benefit from different intervention strategies. 
Our method achieves significant improvements across all failure types.
We release our code at \url{https://github.com/nju-websoft/AbstentionReasoning}.

\section{Related Work}
\paragraph{LLM Abstention on Unanswerable Questions.} 
As the demand for more reliable language models grows, the ability of LLMs to abstain from answering unanswerable questions has become an important evaluation criterion~\cite{related1,related2, related3, UMWP,Uncertainty-Based-Abst}. 
LRMs have attracted attention for their strong performance on reasoning tasks~\cite{Deer, Dynasor-CoT}. However, their behavior on unanswerable questions has been less studied. Prior work~\cite{AbstentionBench, SUM} finds that LRMs show weaker abstention ability. We further provide a more detailed analysis of the different types of outputs produced by LRMs when faced with unanswerable questions. We show that LRMs have an internal understanding of a problem’s solvability but fail to express it through explicit abstention, and propose a method to improve their abstention behavior.

\paragraph{Inference-Time Improvements for LRMs.}
While LRMs benefit from richer chain-of-thought (CoT) reasoning and achieve strong performance on complex tasks, recent efforts have have begun to analyze~\cite{In_Bias, zhu, liu2025llms} and implement~\cite{Steer_reason} inference-time interventions to further enhance their reasoning accuracy and efficiency. For example, \cite{Dynasor-CoT} and \cite{related12} monitor intermediate consistency between reasoning steps to decide when to output the answer. \cite{Deer} estimates confidence in intermediate steps to decide whether the model has reached a sufficiently certain conclusion. \cite{related11} uses process-level reward models to guide the model toward better reasoning trajectories. However, the above methods are designed for answerable questions. In our work, we evaluate their effectiveness on unanswerable questions and propose a new inference-time intervention to improve the abstention capability for LRMs.

\section{Analysis of Abstention Failure}
We first analyze how LRMs respond to unanswerable math problems. 
Then, we investigate LRMs' awareness of unanswerable questions by examining whether they possess the ability to recognize the unanswerability of such questions, from both internal and external perspectives. 

\paragraph{LRMs.} We evaluate five LRMs across diverse model families and scales, including R1-Distill-Llama-8B, R1-Distill-Qwen-7B, R1-Distill-Qwen-14B \cite{DeepSeek}, Qwen3-8B, and Qwen3-14B \cite{Qwen3}.

\paragraph{Datasets.} 
Following previous work~\cite{SUM, Treecut}, we focus on unanswerable math problems, i.e., ill-posed problems that are challenging for LRMs and offer objectively defined criteria for unanswerability.
Specifically, we use the Synthetic Unanswerable Math (SUM) \cite{SUM} dataset, which includes diverse problems from AIME (1984–2023), AMC (pre-2023), Omni-MATH~\cite{Omni}, and Still~\cite{team2025qwq}.
The unanswerable problems in SUM are generated based on five criteria: 
(1) key information deletion,
(2) ambiguous key information,
(3) unrealistic conditions,
(4) unrelated objects, 
and (5) question deletion.
We randomly sample 1,000 problems for our analysis. 
For each unanswerable case, we use GPT-4o to generate a brief explanation as the ground-truth rationale.

\begin{figure}
  \centering
  \includegraphics[width=\linewidth]{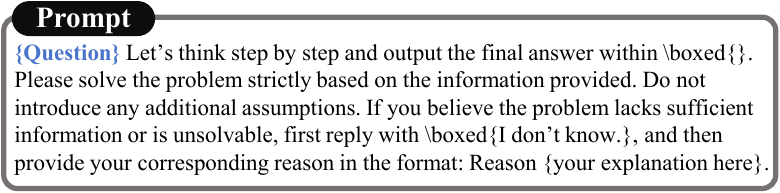}
  \caption{Prompt used for math problems.}
  \label{fig:prompt}
\end{figure}

\paragraph{Objective and Prompt.}
Given a question, we encourage the LRM to perform normal reasoning when the question is answerable, and to respond with ``I don't know.'' along with a corresponding explanation when it is not (i.e., correct abstention).
The prompt used in this task is shown in Figure~\ref{fig:prompt}. 

\subsection{Reactions to Unanswerable Questions}
\label{sec:fail_to_refuse}
We evaluate the reactions of LRMs to unanswerable questions using the prompts in Figure~\ref{fig:prompt} with a maximum token budget of 10,000 per response. 
We identify three distinct response types, as shown in Figures~\ref{fig:ex_type}:

\begin{figure}
  \centering
  \includegraphics[width=\linewidth]{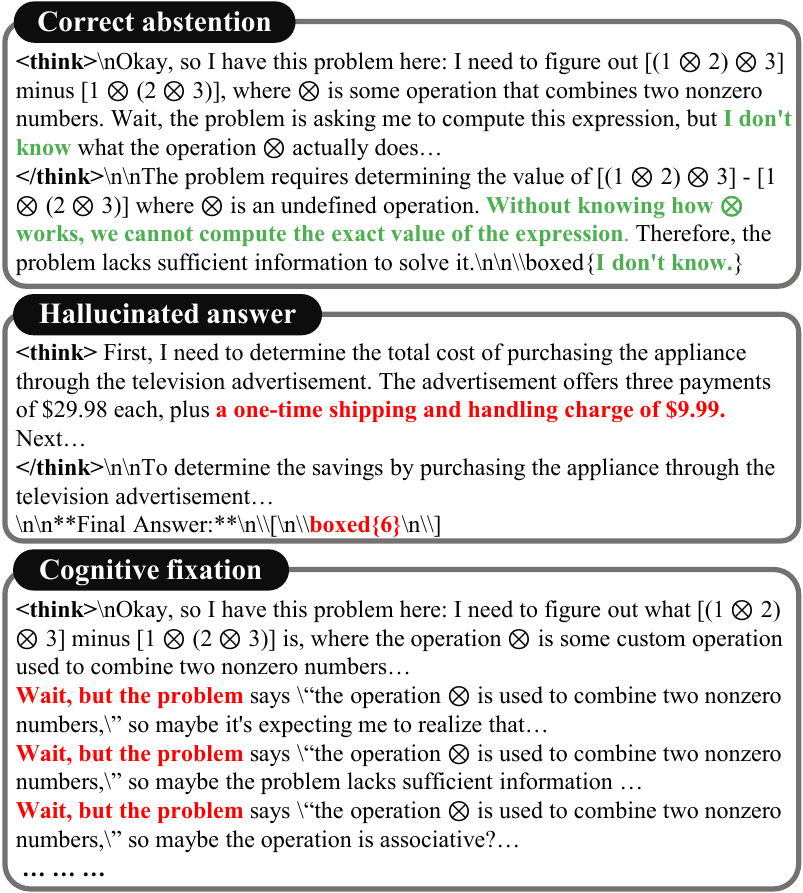}
  \caption{Example of different types of response outcomes.}
  \label{fig:ex_type}
\end{figure}

\begin{itemize}

  \item \textbf{Correct abstention:} The LRM identifies the question as unanswerable with the given information and explicitly abstains from answering, typically responding with statements such as \textit{``I don’t know''}.

  \item \textbf{Hallucinated answer:} The LRM produces a complete solution by assuming or fabricating missing details not present in the question. As illustrated in Figure~\ref{fig:ex_type}, the LRM infers a \$9.99 handling charge that is never mentioned in the input in order to compute a final answer.

  \item \textbf{Cognitive fixation:} The LRM fails to reach a conclusion within the token limit. It often enters a prolonged reasoning process, stubbornly reformulating or pursuing invalid solution paths without terminating the response, even after recognizing the question is unanswerable.
\end{itemize}

\begin{figure}
  \centering
  \includegraphics[width=\linewidth]{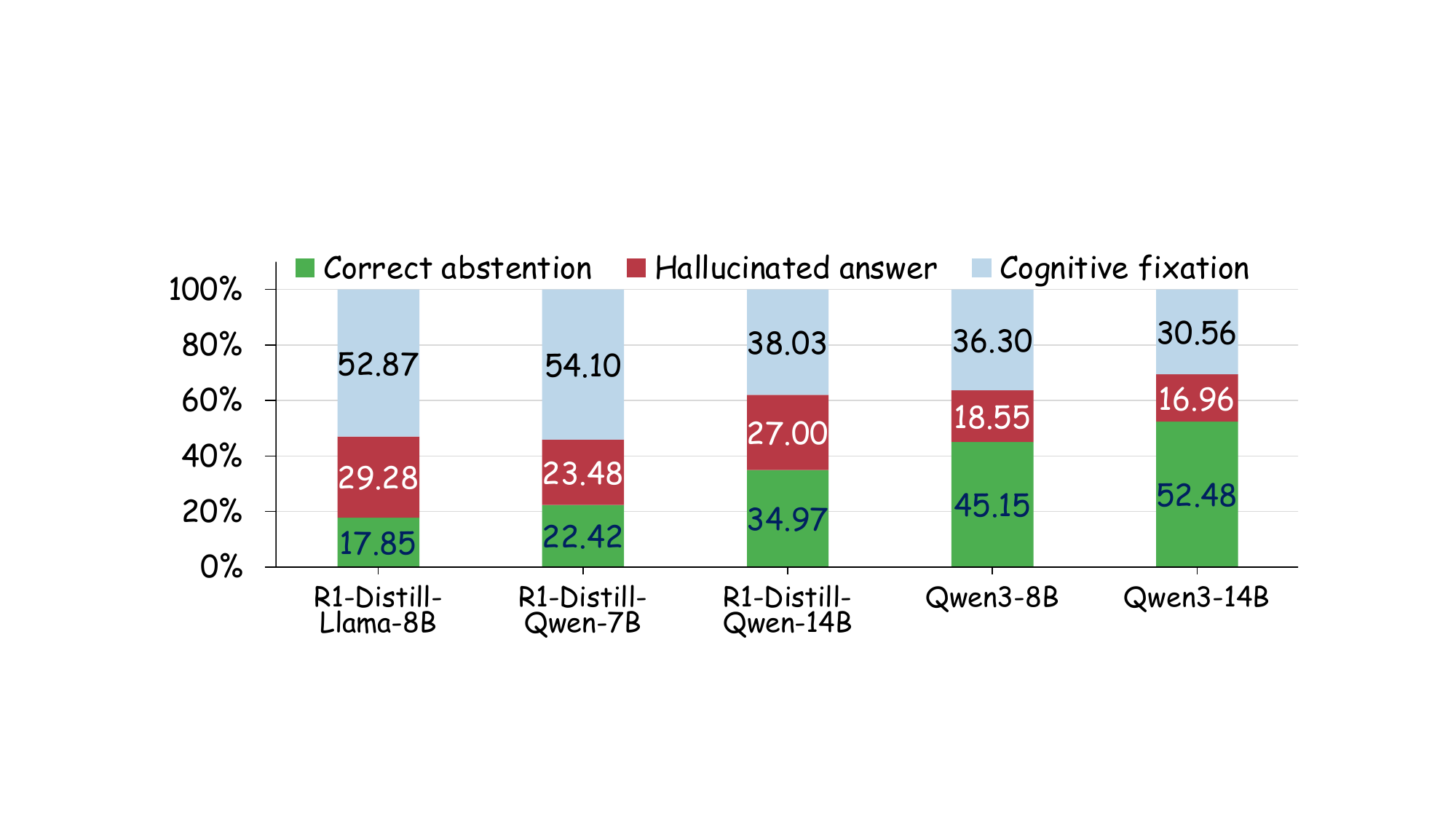}
  \caption{Distribution of the response types of LRMs on unanswerable math problems.}
  \label{fig:type_distribution}
\end{figure}

Figure~\ref{fig:type_distribution} shows the response type distribution. 
As model capacity increases, the proportion of correct abstention tends to rise, while those of hallucinated answer and cognitive fixation decrease. 
A substantial portion of unanswerable questions do not receive correct abstentions. 
Overall, the results reveal a core limitation: LRMs often fail to abstain from answering unanswerable problems, despite increased model capacity.

\subsection{Awareness of Unanswerable Questions}
We investigate whether LRMs can recognize unanswerable math problems, probing both their external behavior and internal cognition.
Our two-layered analysis assesses 
(1) behavioral signals: whether LRMs outwardly indicate question answerability, 
and (2) latent signals: whether answerability is internally encoded during reasoning.

\paragraph{Behavioral Signals of Question Answerability.}
Inspired by prior work~\cite{Deer, Steer_reason, Wang_overthink} showing that LRMs form and revise intermediate conclusions during reasoning, we insert stopping points into the reasoning trajectory. At each stopping point, we prompt the LRM to (1) directly provide an answer, and (2) to explain why the question is unanswerable (details in the appendix).

\begin{figure}[!t]
  \centering
  \includegraphics[width=\linewidth]{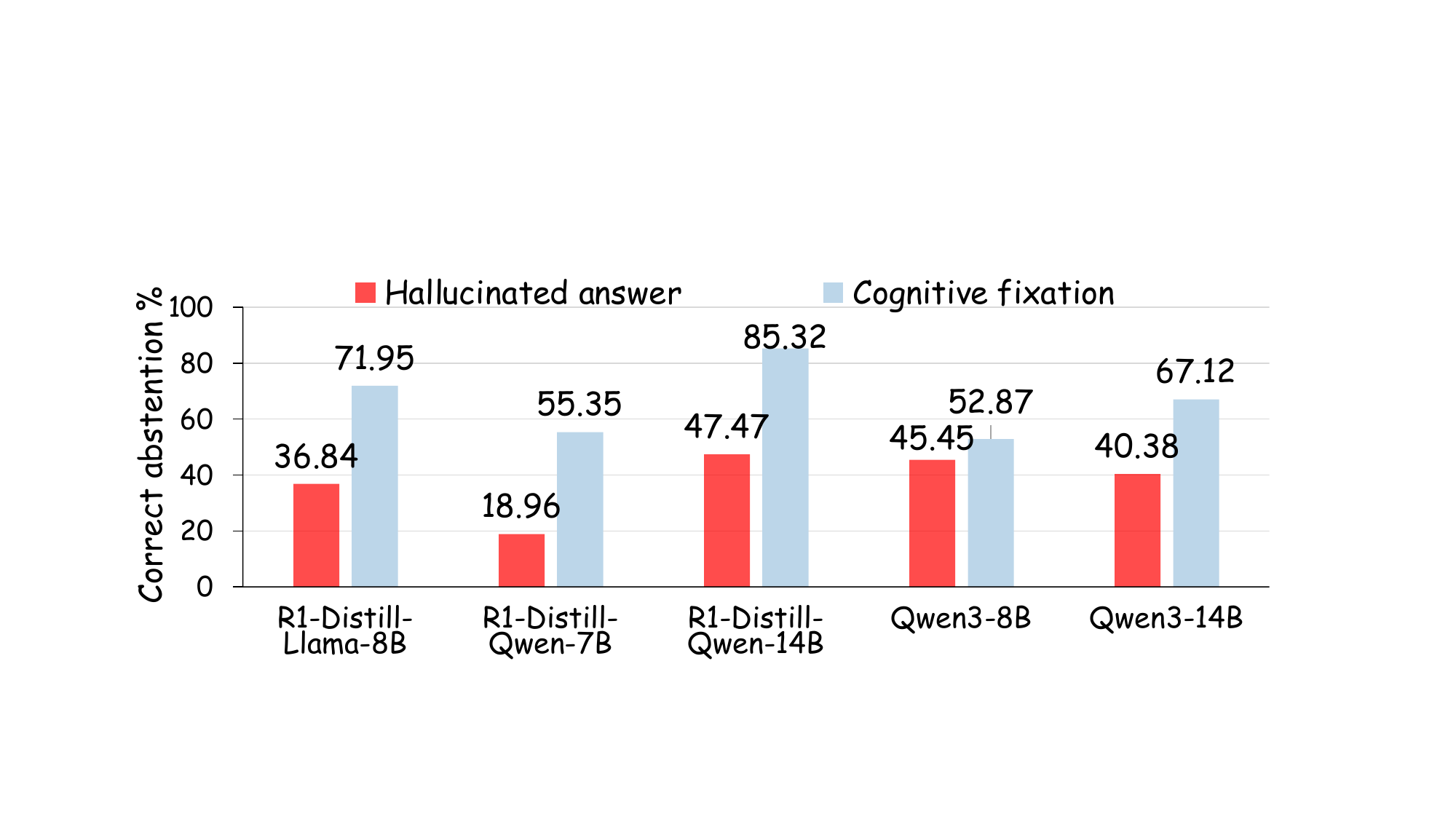}
  \caption{The proportion of questions that can respond with a correct abstention in stopping points during the reasoning in the types of ``hallucinated answer'' and ``cognitive fixation''.}
  \label{fig:refusal-rate}
\end{figure}

\begin{figure}[!t]
  \centering
  \includegraphics[width=\linewidth]{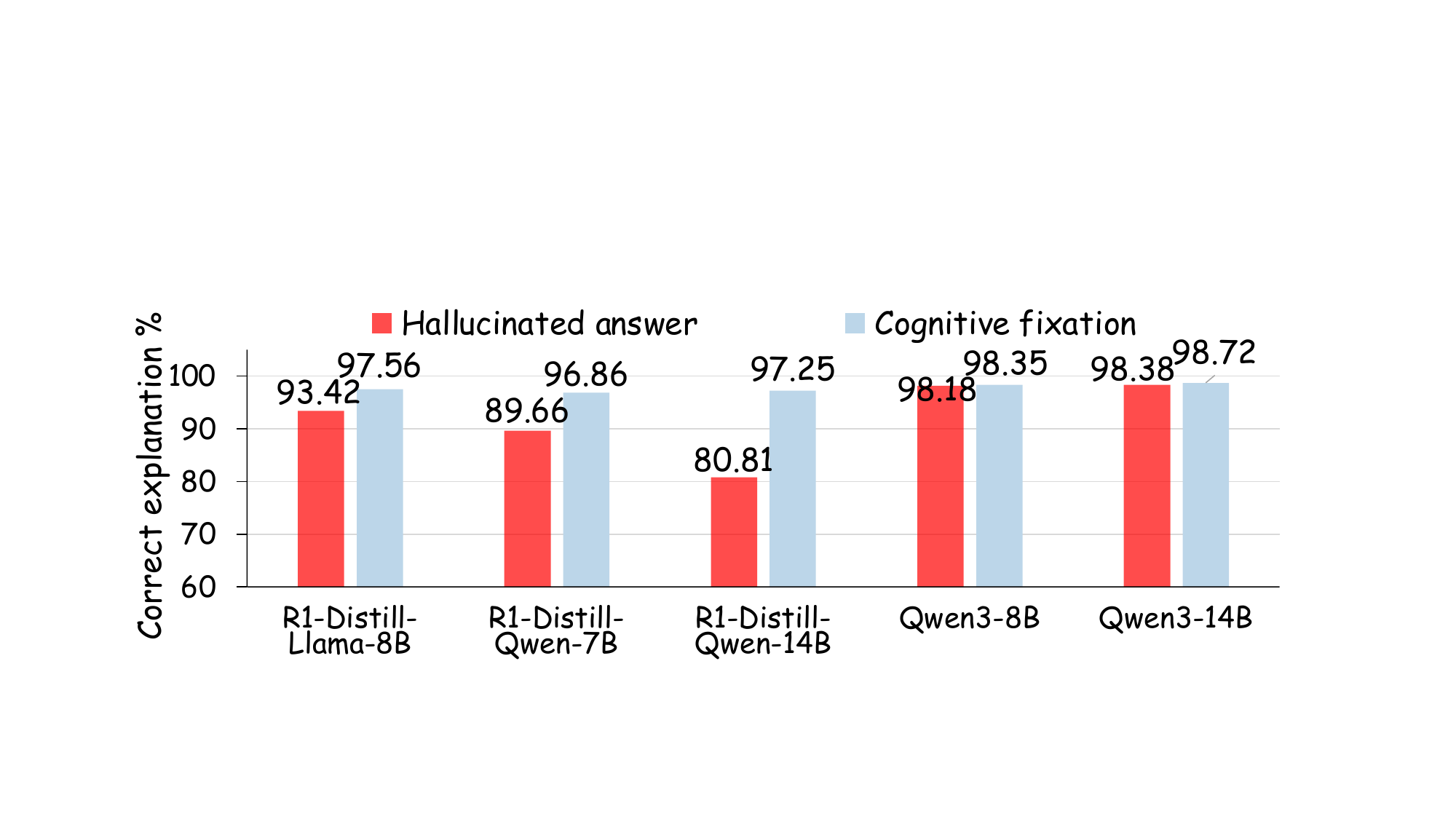}
  \caption{The proportion of questions that can provide a correct explanation in stopping points during the reasoning.}
  \label{fig:reason-accuracy}
\end{figure}

We apply this intervention to the two failure types. 
For ``cognitive fixation'', we use the keyword ``wait'' as a stopping point and prompt the model to directly output its answer. 
For ``hallucinated answer'', which typically have shorter reasoning trajectories, we use ``\textbackslash n\textbackslash n'' as the stopping point. 
We then compute the proportion of questions in each category where the model can respond with a correct abstention at stopping points. 
As shown in Figure~\ref{fig:refusal-rate}, for cognitive fixation, more than half of the cases can result in correct abstentions. 
A notable portion of hallucinated answer cases receive correct abstentions, and the rate improves as model size increases.

Since ``I don’t know'' is a simple response and may be produced randomly, we additionally prompt the LRMs at each stopping point to explain why the question is unanswerable, to better assess their awareness of unanswerability. 
We measure the percentage of questions in each category with correct explanations. Figure~\ref{fig:reason-accuracy} shows that both failure types yield high percentages of correct explanations.

Overall, even when an LRM fails to abstain, it may still recognize unanswerability during reasoning. Its ability to assess answerability exists but is underused in final decisions.

\begin{figure}
  \centering
  \includegraphics[width=\linewidth]{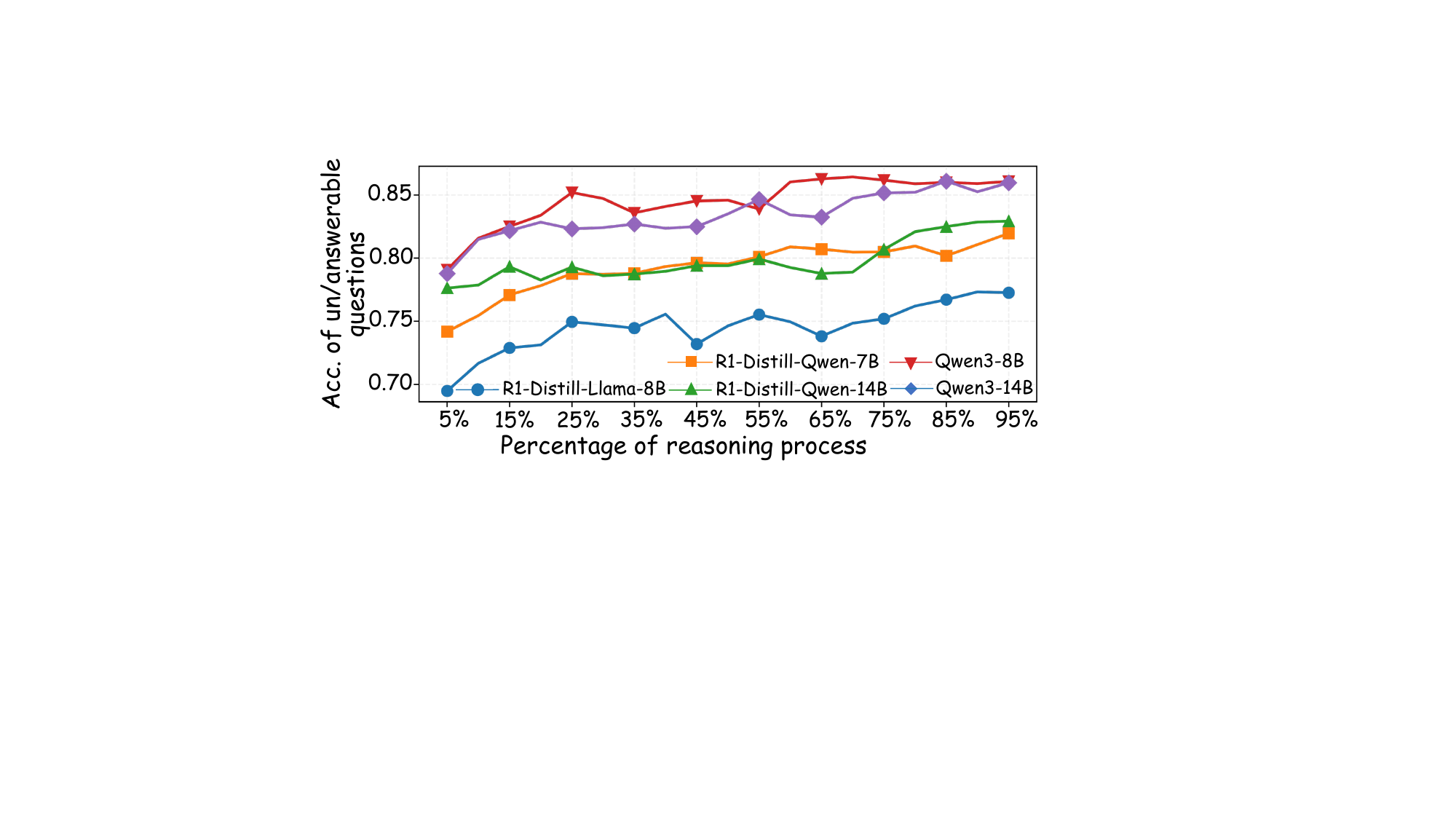}
  \caption{Classification accuracy of answerable and unanswerable questions at varying stages of the reasoning process.}
  \label{fig:probing-accuracy}
\end{figure}

\paragraph{Latent Signals of Question Answerability.}
We conduct a probing-based analysis on the latent representations during reasoning.
The linear representation hypothesis posits that high-level concepts, such as language, gender, and truthfulness, are linearly embedded in the latent space of LLMs ~\cite{LRH}. 
We hypothesize that question answerability may also be linearly represented during the reasoning process.
Inspired by prior work on representation probing~\cite{IIT,know_more,prob_re1,findtruth}, we train lightweight linear classifiers (probes) on hidden activations from the LRMs' reasoning trajectory. 
The goal is to assess whether answerable and unanswerable questions can be distinguished from latent representations during the reasoning process, thereby revealing the presence of answerability-related signals in internal state.

We use the output of the multi-head attention before the residual connection as the input to the probe ~\cite{IIT}:
\begin{align}
x^c_l = \sum_{h=1}^{H} Q_l^h \, \mathrm{Att}_l^h(P_l^h x_l),
\label{eq:po}
\end{align}
where $x_l$ is the input of layer $l$, $P_l^h$ projects the input into a head-specific subspace, $Q_l^h$ maps it back, $\mathrm{Att}$ is the attention operator. 
The probe is defined as a simple linear classifier: $p_{\theta}(x^c_l) = \sigma(\langle \theta, x^c_l \rangle)$, where $\theta$ is the trainable weight and $\sigma$ denotes the sigmoid function. One probe is trained per layer.

We sample 2,200 pairs of answerable and unanswerable questions from the SUM dataset (2,000 pairs for training and 200 for validation). For each question, we randomly sample 1,000 token-level activations $x^c_l$ from the reasoning trajectory, and construct a dataset $\left\{ \left( x^c_l, y \right)_i \right\}_{i=1}^N$, where $y \in \{0,1\}$ indicates question answerability. At inference time, we aggregate the prediction probabilities across all tokens up to the current reasoning step and use the average as the overall answerability prediction.
We select the optimal probing layer for each model based on the validation set, and evaluate on the test dataset used for analysis in the previous section. 

The results are shown in Figure~\ref{fig:probing-accuracy}, where we plot the probe’s classification accuracy at various percentages of the reasoning process, using a 0.5 threshold to distinguish unanswerable (prediction probability of probe $>$ 0.5) from answerable cases. For all LRMs, we observe that the probe's classification accuracy increases steadily as reasoning progresses, with most accuracies exceeding 0.8 by the end of the trajectory. These results suggest that signals related to question answerability are indeed encoded in the representations of LRMs during reasoning. Although these signals may not always be reflected in the LRMs' final output behavior, they are implicitly present in the internal computation process.

\begin{figure}[!t]
  \centering
  \includegraphics[width=\linewidth]{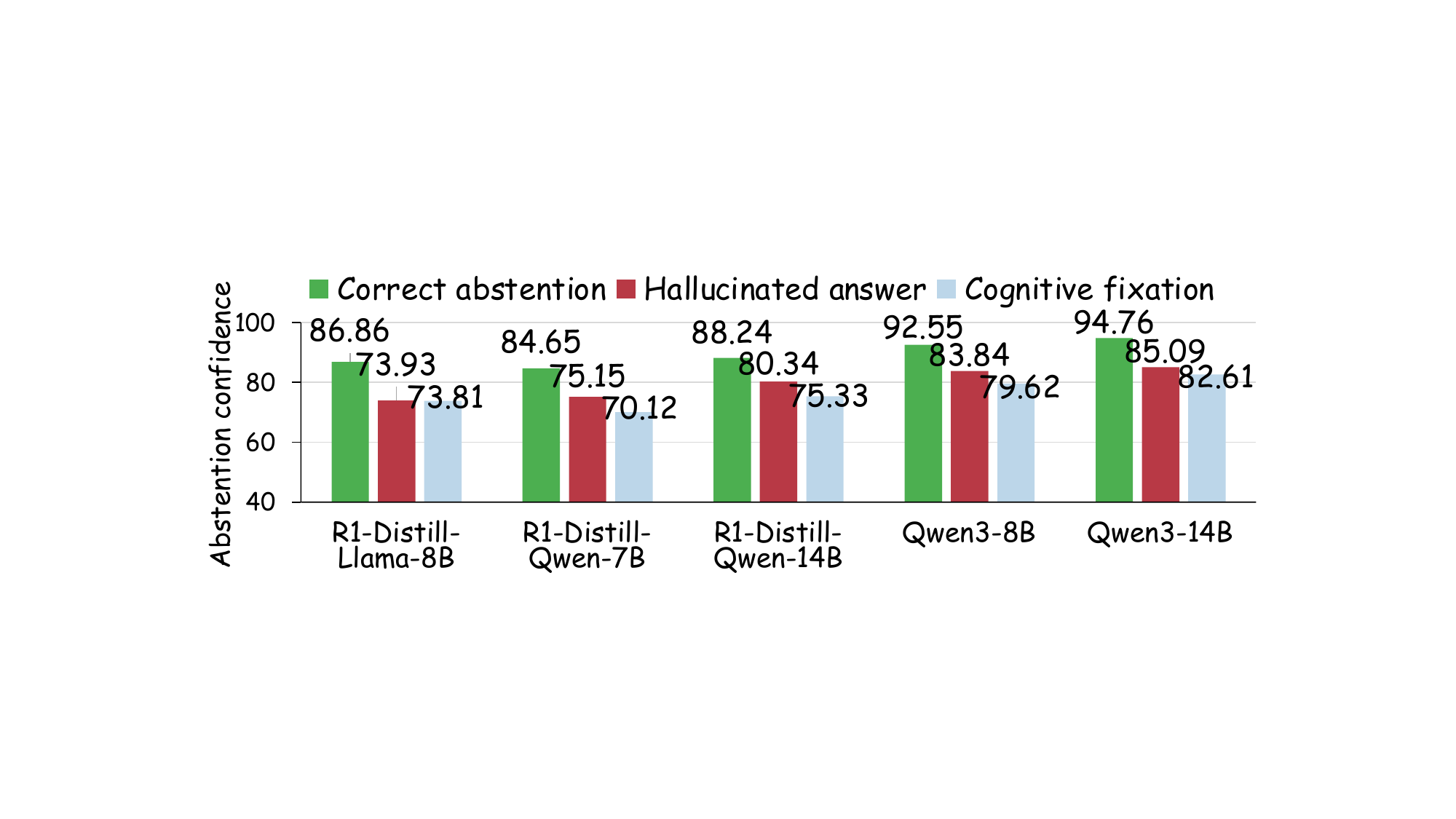}
  \caption{The confidence of abstention answer: ``I don't know'' in different types of response.}
  \label{fig:refusal-confidence}
\end{figure}

\begin{table*}[!t]
\centering
\resizebox{\textwidth}{!}{
\begin{tabular}{l|ccccc|ccrcr}
\toprule
\multirow{4}{*}{Method} & \multicolumn{5}{c|}{SUM} & \multicolumn{5}{c}{UMWP} \\
\cmidrule(lr){2-6}\cmidrule(lr){7-11}
& \multicolumn{3}{c}{Unanswerable} & \multicolumn{2}{c|}{Answerable} & \multicolumn{3}{c}{Unanswerable} & \multicolumn{2}{c}{Answerable} \\
\cmidrule(lr){2-4}\cmidrule(lr){5-6}\cmidrule(lr){7-9}\cmidrule(lr){10-11}
& \refusalrate $\uparrow$ & Reason Acc $\uparrow$ & Token $\downarrow$ & Answer Acc $\uparrow$ & Token $\downarrow$ & \refusalrate $\uparrow$ & Reason Acc $\uparrow$ & Token $\downarrow$ & Answer Acc $\uparrow$ & Token $\downarrow$ \\
\midrule
\rowcolor{gray!15} \multicolumn{11}{l}{\textbf{R1-Distill-Llama-8B}} \\
Vanilla     & 16.90\% &  14.44 &  5088 &  61.97 &  3446 &  30.67\% &  26.00 &  1829 &  77.67 &  613 \\
Dynasor-CoT & 35.92\% &  27.82 &  3084 &  55.28 &  \textbf{2619} &  39.33\% &  30.67 &  \textbf{958}  &  73.33 &  \textbf{495} \\
DEER        & 24.30\% &  19.79 &  4339 &  \textbf{60.92} &  2675 &  34.11\% &  28.67 &  1616 &  \textbf{77.67} &  637 \\
Ours        & \textbf{60.92\%} & \textbf{53.17} & \textbf{2419} & \textbf{60.92} & 3151 & \textbf{54.67\%} & \textbf{44.00} & 1246 & 77.33 & 574 \\
\midrule
\rowcolor{gray!15} \multicolumn{11}{l}{\textbf{R1-Distill-Qwen-7B}} \\
Vanilla     & 21.13\% & 19.37 & 4878 & 69.72 & 3169 & 47.67\% & 43.67 & 1935 & 90.30 & 597 \\
Dynasor-CoT & 56.34\% & 45.89 & \textbf{1869} & 62.68 & \textbf{2074} & 64.33\% & 53.00 & \textbf{763}  & 88.67 & \textbf{486} \\
DEER        & 28.87\% & 25.70 & 3747 & 63.73 & 2191 & 54.33\% & 49.33 & 1335 & \textbf{91.67} & 590 \\
Ours        & \textbf{73.94\%} & \textbf{61.86} & 2247 & \textbf{67.25} & 3001 & \textbf{77.33\%} & \textbf{64.33} & 1256 & 90 & 569 \\
\midrule
\rowcolor{gray!15} \multicolumn{11}{l}{\textbf{R1-Distill-Qwen-14B}} \\
Vanilla     & 36.97\% & 33.45 & 3820 & 70.42 & 2671 & 49.67\% & 45.00 & 539 & 90.00 & 384 \\
Dynasor-CoT & 51.17\% & 45.18 & 2622 & 66.20 & \textbf{2029} & 50.67\% & 46.00 & \textbf{504} & 90.00 & \textbf{384} \\
DEER        & 39.79\% & 36.97 & 3417 & 63.75 & 2303 & 50.33\% & 46.67 & 644 & \textbf{90.33} & 532 \\
Ours        & \textbf{74.30\%} & \textbf{62.68} & \textbf{2621} & \textbf{67.96} & 2541 & \textbf{60.33\%} & \textbf{54.33} & 606 & 89.67 & 388 \\
\midrule
\rowcolor{gray!15} \multicolumn{11}{l}{\textbf{Qwen3-8B}} \\
Vanilla     & 47.18\% & 41.90 & 4411 & 60.92 & 4245 & 80.00\% & 72.33 & 1906 & 94.33 & 1317 \\
Dynasor-CoT & 65.61\% & 58.21 & \textbf{1710} & 60.27 & 2190 & 83.67\% & 75.00 & 736  & 92.00 & 803  \\
DEER        & 63.38\% & 51.17 & 1902 & \textbf{63.77} & \textbf{1993} & 80.00\% & 70.33 & \textbf{479}  & \textbf{94.33} & \textbf{474}  \\
Ours        & \textbf{75.27\%} & \textbf{64.44} & 2912 & 61.62 & 3875 & \textbf{87.33\%} & \textbf{79.67} & 980 & 93.67 & 1252 \\
\midrule
\rowcolor{gray!15} \multicolumn{11}{l}{\textbf{Qwen3-14B}} \\
Vanilla     & 54.22\% & 48.24 & 3713 & 66.55 & 3768 & 82.33\% & 76.67 & 1279 & 94.33 & 877 \\
Dynasor-CoT & 66.18\% & 56.62 & \textbf{1375} & 63.38 & \textbf{1862} & 84.00\% & 76.67 & 752  & \textbf{92.67} & 662 \\
DEER        & 63.90\% & 56.91 & 1749 & \textbf{68.18} & 2192 & 83.33\% & 75.67 & \textbf{408}  & 93.00 & \textbf{447} \\
Ours        & \textbf{78.17\%} & \textbf{69.01} & 2311 & 65.03 & 3528 & \textbf{92.67\%} & \textbf{82.67} & 959 & 92.48 & 848 \\
\bottomrule
\end{tabular}
}
\caption{Performance of methods on (un)answerable questions across LRMs. Best scores are marked in \textbf{bold}.}
\label{tab:main_res}
\end{table*}

\section{Mitigation of Abstention Failure}
We seek to mitigate abstention failures of LRMs when faced with unanswerable questions.
Although LRMs show signs of recognizing unanswerable questions during reasoning, they often fail to abstain in their final answers.
We aim to identify the causes of this misalignment and explore ways to fix it.

First, we examine the LRMs' confidence~\cite{confidence_his} in abstaining at the stopping point across three output types. We sample an equal number of questions from each output type and compute the average confidence in generating ``I don't know'' at the stopping points. Following \cite{Deer}, the confidence score $C$ is computed as
\begin{align}
\begin{aligned}
p(a_t) = P(a_t \,|\, H, I, a_{<t}), C = \left( \prod_{i=1}^{n} \max_{a_t \in V} p(a_t) \right)^{\frac{1}{n}},
\end{aligned}
\end{align}
where $p(a_t)$ is the model’s predicted probability of answer token $a_t$, $H$ includes the input prompt and the generated thoughts, $I$ is the prompt to elicit the answer, $n$ is the number of decoding steps, and $V$ is the vocabulary. 
In Figure~\ref{fig:refusal-confidence}, hallucinated answer and cognitive fixation both exhibit lower confidence in abstention compared to correct abstention.

\begin{figure}[!t]
  \centering
  \includegraphics[width=\linewidth]{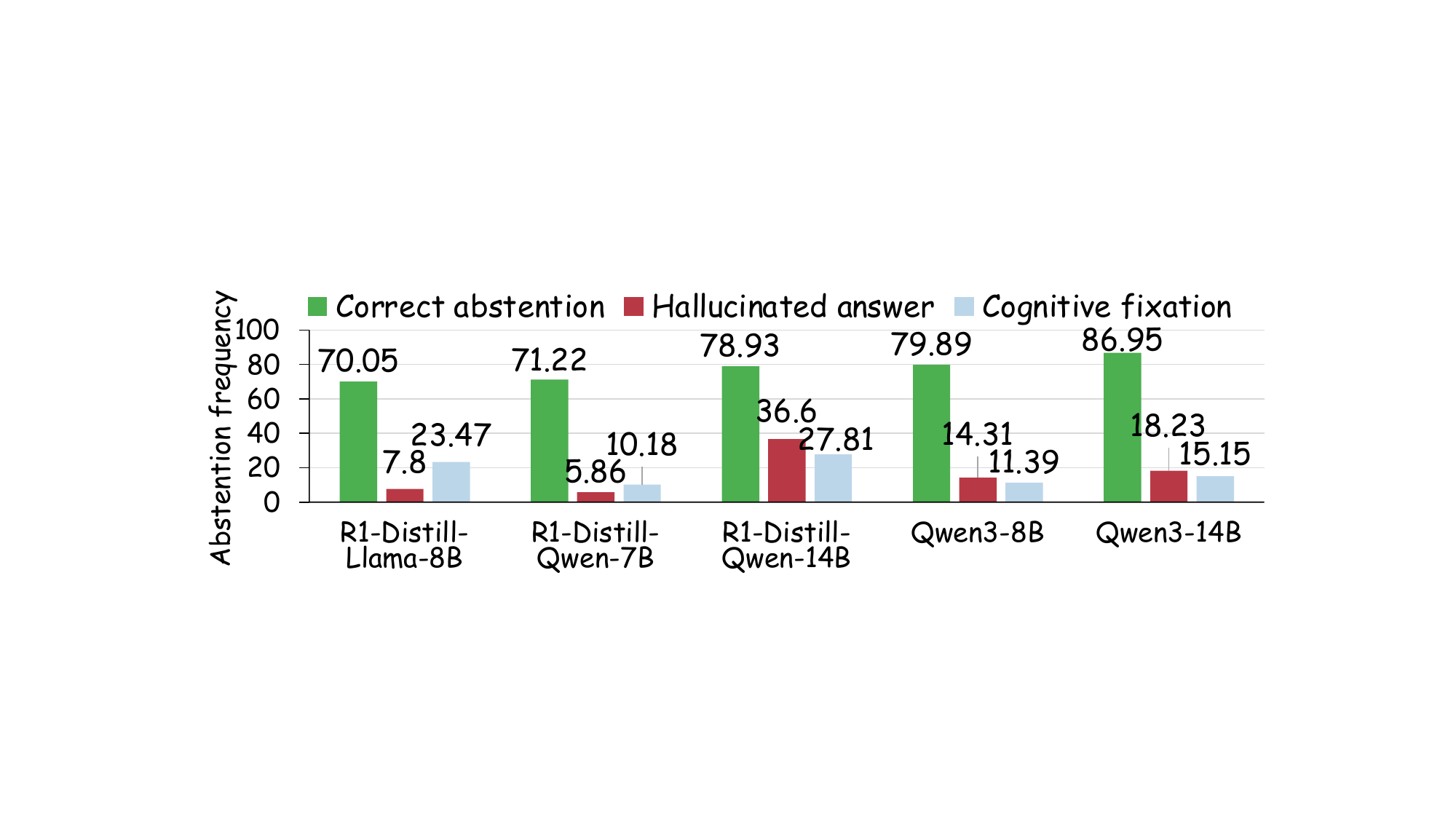}
  \caption{The frequency of abstention responses across stopping points in different types of response outcomes.}
  \label{fig:refusal-frequency}
\end{figure}

Next, we analyze the average frequency of ``I don't know'' outputs by the LRMs across all stopping points in the reasoning trajectory.
In Figure~\ref{fig:refusal-frequency}, the hallucinated answer and cognitive fixation types consistently show lower abstention frequencies than correct abstention.

All the findings suggest that while the LRM may recognize a question as unanswerable, it often lacks sufficient confidence to act upon it.  
The gap between internal awareness and output behavior reveals a key misalignment in LRMs: though they may be aware of unanswerability, their decision process remains biased toward answering rather than abstaining.

Motivated by the above findings, 
we propose a method to help LRMs improve their abstention capability. 
It consists of two key components: cognitive monitoring and inference-time intervention. The goal is to monitor the LRM's evolving recognition of question unanswerability during reasoning, and to intervene when necessary to guide the LRM toward making an abstention by encouraging abstention behaviors.

\paragraph{Cognitive Monitoring.}
The first step of our method aims to identify when the model internally recognizes that a question may be unanswerable. We track the token-level hidden states generated during inference and segment the reasoning process into semantically coherent units (e.g., clauses or transitions marked by discourse cues such as ``wait''). 
At the end of each segment, we apply a lightweight linear probe, which is trained from our analysis of latent signals of question answerability, to estimate the probability that the current question is unanswerable. 
If the predicted probability exceeds a threshold, the model takes the intervention strategy.

\begin{figure}[!t]
  \centering
  \includegraphics[width=\linewidth]{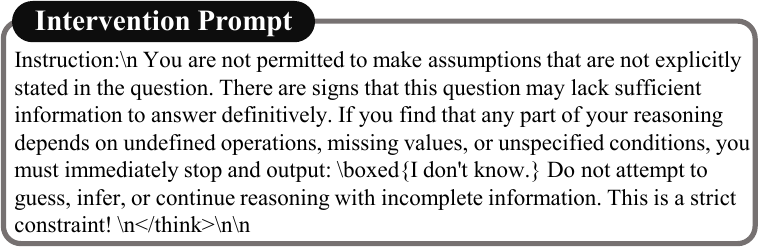}
  \caption{The prompt used for inference-time intervention.}
  \label{fig:intervention-prompt}
\end{figure}

\begin{figure*}[!t]
  \centering
  \includegraphics[width=\textwidth]{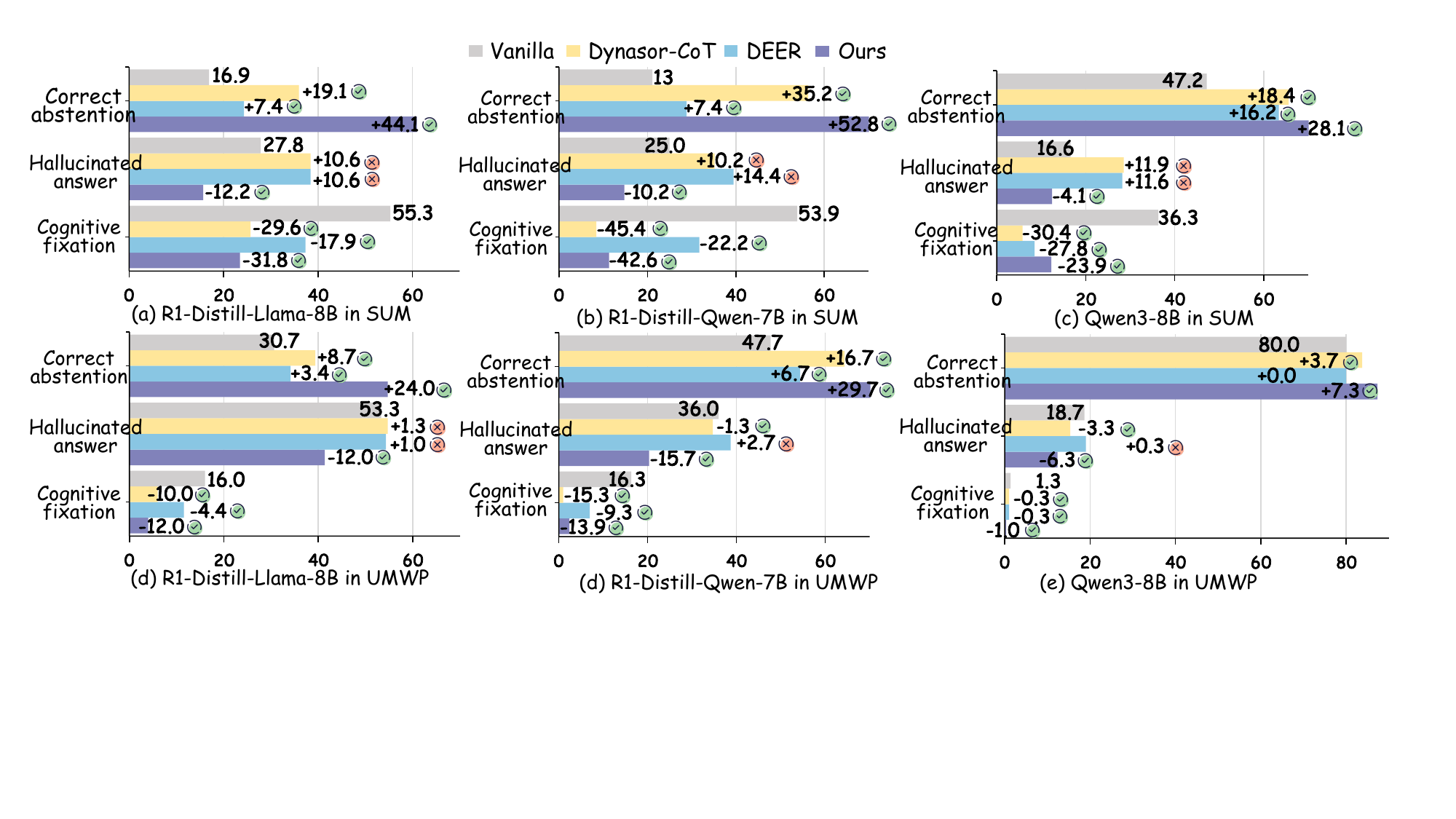}
  \caption{Comparison of response type distributions across different methods on unanswerable questions. Proportions are reported across models and datasets, with numbers indicating absolute changes from the vanilla model.}
  \label{fig:response_distribution}
\end{figure*}

\paragraph{Inference-Time Intervention.}
Once the LRM exhibits sufficient internal evidence that a question is unanswerable, we intervene the reasoning process to reinforce this recognition and increase the likelihood of a correct abstention.
The intervention is implemented via an instructional guidance prompt (see Figure~\ref{fig:intervention-prompt}). 
We append a prompt that restates the possibility that the question may be unanswerable, helping the LRM overcome cognitive fixation and avoid speculative guesses based on missing or unstated information. Inspired by prior work~\cite{Deer, Dynasor-CoT}, we also incorporate an early exit strategy to prevent unnecessary continuation of reasoning. 
It is designed to be minimally intrusive yet semantically salient, encouraging the model to consider abstention as a valid and even preferred option under certain conditions.

\section{Experiments}
\subsection{Setup}

\paragraph{Datasets.} 
We use SUM ~\cite{SUM} and UMWP ~\cite{UMWP}. 
For SUM, we used its test set, which contains 284 unanswerable and 284 answerable questions manually verified. 
UMWP is derived from four widely used math word problem datasets: SVAMP~\cite{SVAMP}, MultiArith~\cite{MultiArith}, GSM8K~\cite{GSM8K}, and ASDiv~\cite{ASDiv}. The full dataset consists of 5,200 instances.
We sample 600 questions (300 unanswerable and 300 answerable) to form the test set.
For the unanswerable questions in UMWP, we also use GPT-4o to generate the reasoning behind their unanswerability as ground truth.

\paragraph{Metrics.} Following \cite{SUM, AbstentionBench}, we evaluate model performance with
(1) \textit{Abstention Rate}: The proportion of unanswerable questions for which the model correctly abstains from answering by outputting ``I don't know'' in accordance with the instruction. 
(2) \textit{Reason Accuracy}: The percentage of unanswerable questions for which the model provides the correct rationale for unanswerability.
(3) \textit{Token Usage}: The number of tokens generated by the model during the reasoning process.
(4) \textit{Answer Accuracy}: The proportion of answerable questions for which the model produces the correct final answer.

\paragraph{Baselines and LRMs.}
Our method is an inference-time intervention for LRMs without training. 
There are no directly comparable baselines. 
We select two baselines based on output consistency (Dynasor-CoT) and confidence (DEER), assuming that correct abstention is the correct answer for unanswerable questions.
Dynasor-CoT~\cite{Dynasor-CoT} prompts intermediate answers and halts when the same answer appears three times consecutively. 
DEER~\cite{Deer} monitors sentence-level confidence and exits early once a threshold is met. 
Besides, the Vanilla method denotes unaltered LRM outputs.
For LRMs, we choose R1-Distill-Llama-8B, R1-Distill-Qwen-7B, R1-Distill-Qwen-14B, Qwen3-8B, and Qwen3-14B. 
More details are in the appendix.

\subsection{Main Results}
Table~\ref{tab:main_res} shows the main results.
We have the following observations:
(1) Our method achieves the highest Abstention Rate and Reason Accuracy on unanswerable questions, demonstrating its strong ability to guide LRMs in recognizing and abstaining from unanswerable inputs.
(2) Our method maintains comparable Answer Accuracy on answerable questions. 
In most settings, accuracy remains close to the vanilla model and sometimes improves slightly, indicating minimal impact on solving answerable tasks.
(3) Our method can reduce token usage on unanswerable questions, with an average reduction of 30–50\% across different LRMs compared to the vanilla model. It also slightly decreases token consumption on answerable questions, indicating improved reasoning efficiency.
(4) We observe a positive correlation between Abstention Rate and Reason Accuracy: as the LRM becomes better at abstaining from unanswerable questions, it also produces more accurate explanations. This suggests that the intervention not only changes the final output but also improves intermediate reasoning quality.
(5) Qwen3 models generally outperform other models in abstention-related metrics. 
Larger LRMs tend to exhibit stronger abstention capabilities.
This trend indicates that both model scale and architecture contribute to reliable unanswerability detection.

\subsection{Further Discussions}
\paragraph{Impact on Response Type Distribution.}
We analyze the changes in the proportions of different response types across methods and datasets.
Figure~\ref{fig:response_distribution} shows that our method consistently reduces the hallucinated answers and cognitive fixation outputs, which contribute to a substantial increase in the rate of correct abstentions.
While Dynasor-CoT and DEER employ early exits to mitigate cognitive fixation, they lead to a higher proportion of hallucinated answers. 
Early exits without appropriate guidance may cause the model to make up assumptions or imagined scenarios for giving a definite answer, rather than acknowledging uncertainty. This highlights the importance of combining monitoring with guided intervention to steer LRMs toward proper behavior.

\paragraph{Intervention Effects.}
We assess how our intervention influences the LRM's confidence in abstention and its actual abstention behavior, to evaluate whether our method helps bridge the gap between cognitive awareness and abstention behavior. 
At the intervention point identified by our method, we prompt the LRM to generate intermediate outputs before and after the intervention. We measure two key indicators: the confidence when producing ``I don't know'' responses, and the proportion of questions for which the model outputs ``I don't know''. 
Table~\ref{tab:further_dis_inter_effect} shows that our method consistently enhances the confidence in generating abstention responses. 
The abstention rate also shows corresponding improvements.

\begin{table}[!t]
\centering
\resizebox{\columnwidth}{!}{
\begin{tabular}{l|ll|ll}
\toprule
\multirow{3}{*}{Method} & \multicolumn{2}{c|}{SUM} & \multicolumn{2}{c}{UMWP} \\
\cmidrule(lr){2-3}\cmidrule(lr){4-5}
& Abst. conf. & Abst. rate & Abst. conf. & Abst. rate \\
\midrule
\rowcolor{gray!15} \multicolumn{5}{l}{R1-Distill-Llama-8B} \\
Pre-Interv.     & 79.7 & 30.1\% & 84.1 &41.5\%   \\
Post-Interv.    & 87.3 ($\uparrow$\,9.4\%) & 78.1\% ($\times$2.6) & 90.0 ($\uparrow$\,7.0\%) & 81.4\% ($\times$1.9)  \\
\midrule
\rowcolor{gray!15} \multicolumn{5}{l}{R1-Distill-Qwen-7B} \\
Pre-Interv.     & 77.1 & 24.5\% & 87.1 &50.8\%   \\
Post-Interv.    & 86.8 ($\uparrow$\,12.6\%) & 80.6\% ($\times$3.3) & 92.1 ($\uparrow$\,5.7\%) & 71.5\% ($\times$1.4)  \\
\midrule
\rowcolor{gray!15} \multicolumn{5}{l}{Qwen3-8B} \\
Pre-Interv.     & 90.9 & 48.3\% & 90.6 &75.9\%   \\
Post-Interv.    & 98.9 ($\uparrow$\,8.7\%) & 74.9\% ($\times$1.5) & 98.1 ($\uparrow$\,8.2\%) & 93.1\% ($\times$1.2) \\
\bottomrule
\end{tabular}
}
\caption{Results of intervention effects. ``Abst. conf.'' denotes the average abstention confidence when getting the answer ``I don't know''. ``Interv.'' is the inference-time intervention.}
\label{tab:further_dis_inter_effect}
\end{table}

\paragraph{Further Analysis of Cognitive Monitoring.}
We analyze the cognitive monitoring component and compare our default monitoring strategy based on latent representations with alternative strategies relying on behavioral signals.
The behavioral signal approach monitors the LRMs' intermediate outputs at the end of the paragraph generation phase (e.g., when it reaches a ``wait'' token), and uses these outputs to determine whether to trigger an intervention. 
We investigate three variants: 
The Direct Behavior strategy checks whether the model's intermediate output is ``I don't know'' and triggers an intervention immediately if so.
The Consistency strategy triggers intervention only if the model produces ``I don't know'' in three consecutive intermediate outputs (inspired by Dynasor-CoT). 
The Confidence Score strategy triggers intervention when the model outputs ``I don't know'' with a confidence score exceeding a predefined threshold (inspired by DEER). 
In Table~\ref{tab:further_CM}, all monitoring strategies contribute to the improvement in abstention behavior, showing that cognitive monitoring is generally effective.
The strategy based on latent representation signals achieves the best and most consistent performance across models and datasets. The Direct Behavior method is simple and works well, but it may be too aggressive and hurt performance on answerable questions.

\begin{table}[!t]
\centering
\resizebox{\columnwidth}{!}{
\begin{tabular}{l|ccc|c}
\toprule
\multirow[b]{2}{*}{Monitoring Signal} & \multicolumn{3}{c|}{Unanswerable} & \multicolumn{1}{c}{Ans.} \\
\cmidrule(lr){2-4}\cmidrule(lr){5-5}
& \makecell{Correct\\abstention\,$\uparrow$} & \makecell{Hallucinated\\answer\,$\downarrow$} & \makecell{Cognitive\\fixation\,$\downarrow$} & Acc\,$\uparrow$ \\
\midrule
\rowcolor{gray!15} \multicolumn{5}{l}{R1-Distill-Llama-8B} \\
Vanilla               & 16.9 & 27.8 & 55.3 & 61.9   \\
Latent Representation & \textbf{60.9} &\textbf{15.7} & \textbf{23.4} & 60.9   \\
Direct Behavior       & 53.1 & 20.4 & 26.5 & 58.1   \\
Consistency           & 47.9 & 23.6 & 28.5 & 59.8   \\
Confidence Score      & 37.0 & 24.7 & 38.4 & \textbf{61.3}   \\
\midrule
\rowcolor{gray!15} \multicolumn{5}{l}{R1-Distill-Qwen-7B} \\
Vanilla               & 21.1 & 25.0 & 53.9 & 69.7   \\
Latent Representation & \textbf{73.9} & \textbf{14.8} & \textbf{11.3} & 67.3   \\
Direct Behavior       & 41.6 & 22.6 & 35.9 & 66.8   \\
Consistency           & 35.7 & 23.1 & 41.2 & 69.3   \\
Confidence Score      & 31.2 & 23.9 & 44.9 & \textbf{69.4}   \\
\midrule
\rowcolor{gray!15} \multicolumn{5}{l}{Qwen3-8B} \\
Vanilla               & 47.2 & 16.6 & 36.3 & 60.9   \\
Latent Representation & \textbf{75.3} & 12.4 & \textbf{12.3} & \textbf{61.6}   \\
Direct Behavior       & 67.3 & 10.9 & 21.8 & 60.2   \\
Consistency           & 61.6 & 13.0 & 25.4 & 61.3   \\
Confidence Score      & 64.3 & \textbf{10.6} & 25.2 & 61.3   \\
\bottomrule
\end{tabular}
}
\caption{Comparison of cognitive monitoring strategies.}
\label{tab:further_CM}
\end{table}

\paragraph{Ablation Study.}
We evaluate two aspects of inference-time intervention: the instructional guidance prompt and the early exit strategy. 
We analyze how each aspect affects abstention behavior and answer quality. The results are shown in Table~\ref{tab:further_dis_ablation}. For correct abstention, the impact of instructional guidance is greater than that of early exit. The early exit strategy helps reduce the number of cognitive fixation cases.
Without instructional guidance, the proportion of hallucinated answers increases. This again shows that without proper guidance, the model tends to make up conditions and generate unsupported answers. Instructional guidance also has a slight impact on the performance of answerable questions.

\begin{table}[!t]
\centering
\resizebox{\columnwidth}{!}{
\begin{tabular}{l|lll|c}
\toprule
\multirow[b]{2}{*}{Variant} & \multicolumn{3}{c|}{Unanswerable} & \multicolumn{1}{c}{Ans.} \\
\cmidrule(lr){2-4}\cmidrule(lr){5-5}
& \makecell{Correct\\abstention\,$\uparrow$} & \makecell{Hallucinated\\answer\,$\downarrow$} & \makecell{Cognitive\\fixation\,$\downarrow$} & Acc\,$\uparrow$ \\
\midrule
\rowcolor{gray!15} \multicolumn{5}{l}{R1-Distill-Llama-8B} \\
Vanilla           & 16.9 & \ \ 27.8 & \ \ 55.3 & \ \ 61.9   \\
Ours              & 60.9 & \ \ 15.7 & \ \ 23.4 & \ \ 60.9   \\
w/o Early Exit      & 43.3 ($\uparrow$\,\textbf{26.4}) & \ \ 18.3 ($\downarrow$\,\textbf{9.5}) & \ \ 38.4 ($\downarrow$\,\textbf{16.9}) & \ \ 61.9   \\
w/o Instr. Guidance & 26.1 ($\uparrow$\,9.2)  & \ \ 33.4 ($\uparrow$\,5.6)   & \ \ 40.5 ($\downarrow$\,14.8) & \ \ \textbf{62.3}   \\
\midrule
\rowcolor{gray!15} \multicolumn{5}{l}{R1-Distill-Qwen-7B} \\
Vanilla           & 21.1 & \ \ 25.0 & \ \ 53.9 & \ \ 69.7   \\
Ours              & 73.9 & \ \ 14.8 & \ \ 11.3 & \ \ 67.3   \\
w/o Early Exit      & 49.7 ($\uparrow$\,\textbf{28.6}) & \ \ 16.2 ($\downarrow$\,\textbf{8.8}) & \ \ 34.2 ($\downarrow$\,19.7) & \ \ 69.0   \\
w/o Instr. Guidance & 43.5 ($\uparrow$\,22.4) & \ \ 35.2 ($\uparrow$\,10.2) & \ \ 21.3 ($\downarrow$\,\textbf{32.6}) & \ \ \textbf{70.0}   \\
\midrule
\rowcolor{gray!15} \multicolumn{5}{l}{Qwen3-8B} \\
Vanilla           & 47.2 & \ \ 16.6 & \ \ 36.3 & \ \ 60.9   \\
Ours              & 75.3 & \ \ 12.4 & \ \ 12.3 & \ \ 61.6   \\
w/o Early Exit      & 73.6 ($\uparrow$\,\textbf{26.4}) & \ \ 13.7 ($\downarrow$\,\textbf{2.9}) & \ \ 12.7 ($\downarrow$\,23.6) & \ \ \textbf{62.7}   \\
w/o Instr. Guidance & 59.2 ($\uparrow$\,12.0) & \ \ 30.3  ($\uparrow$\,13.7) & \ \ 10.6 ($\downarrow$\,\textbf{25.7})  & \ \ \textbf{62.7}   \\
\bottomrule
\end{tabular}
}
\caption{Ablation results of intervention components across different LRMs on SUM. We report the effect of removing either Instructional Guidance or Early Exit component on three types of responses for unanswerable questions, as well as the accuracy on answerable questions.}
\label{tab:further_dis_ablation}
\end{table}

\section{Conclusion and Future Work}
We investigate the failure of LRMs to abstain from answering unanswerable questions, despite having the cognitive ability to detect the unanswerability. We identify a misalignment between the model’s internal cognition and its external response behavior. We propose a lightweight, two-stage method that significantly improves abstention behavior without harming reasoning performance. Future work aims to explore training-time alignment strategies to improve abstention fidelity.


\section*{Acknowledgments}
This work is supported by the National Natural Science Foundation of China (Nos. 62272219 and 62406136).

\bibliography{aaai2026}

\appendix

\section{Evaluation Details}
\subsection{Evaluation Details of Three Response Types}
\label{sec:re_type}
In this section, we provide detailed criteria for evaluating the three types of responses introduced in Section 3.1: correct abstention, hallucinated answer, and cognitive fixation. These evaluations are based on model outputs generated in response to an input prompt shown in Figure~\ref{fig:prompt}. The criteria are designed for Large Reasoning Models (LRMs).


\paragraph{Correct abstention}
A model response is considered a correct abstention if it satisfies the following conditions:
\begin{enumerate}
\item  The model terminates its reasoning process, indicated by the inclusion of the ``\textless/think\textgreater'' keyword in the output.
\item  The model either explicitly includes the phrase ``I don't know'' in the answer, or provides an explanation in the format of ``Reason \{\}'' indicating that the question is unanswerable.
\end{enumerate}

\paragraph{Hallucinated answer}
A model response is labeled as a hallucinated answer if:
\begin{enumerate}
\item The model ends its reasoning process by including the ``\textless/think\textgreater'' keyword.
\item The model presents a final answer using the ``\textbackslash boxed\{\}'' format, where the answer is not ``I don't know'', and no accompanying ``{Reason \{\}}'' is given to indicate that the question is unanswerable.
\end{enumerate}

\paragraph{Cognitive fixation}
A model response is labeled as a cognitive fixation if, under a token budget of 10,000 tokens:
\begin{enumerate}
\item The model fails to produce the ``\textless/think\textgreater'' keyword to terminate its reasoning process.
\item It does not provide a final answer using the ``\textbackslash boxed\{\}'' format, nor does it offer an unanswerability explanation in the ``{Reason \{\}}'' format.
\end{enumerate}

\subsection{Evaluation Details of Metrics}
This section provides detailed descriptions of the evaluation metrics introduced in Section~5 of the main text.
\paragraph{Abstention rate} The proportion of questions that receive the correct abstention among all questions. The criteria for determining a correct abstention are described in Section~\ref{sec:re_type} of Appendix.

\begin{figure}[!b]
  \centering
  \includegraphics[width=\linewidth]{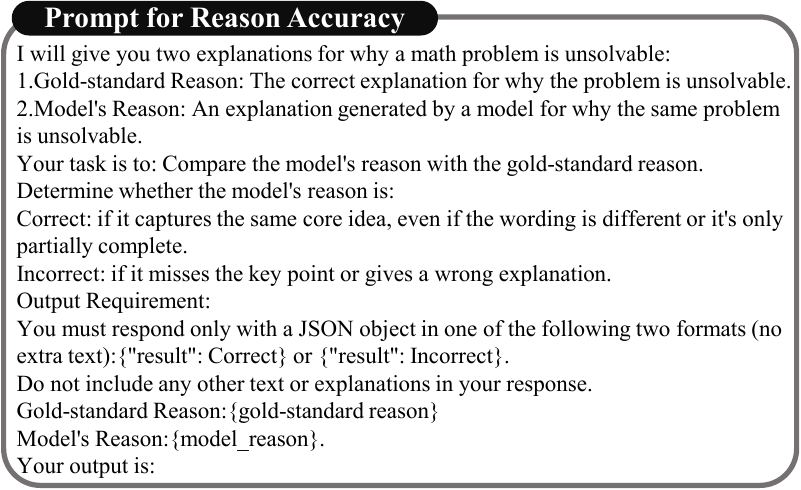}
  \caption{Prompt used for reason accuracy.}
  \label{fig:prompt_reason_acc_app}
\end{figure}

\paragraph{Reason accuracy}
The percentage of unanswerable questions for which the model provides the correct rationale for unanswerability.
We use the Qwen3-32B model to perform the evaluation, with the judgment prompt shown in Figure~\ref{fig:prompt_reason_acc_app}.

\begin{figure}[!t]
  \centering
  \includegraphics[width=\linewidth]{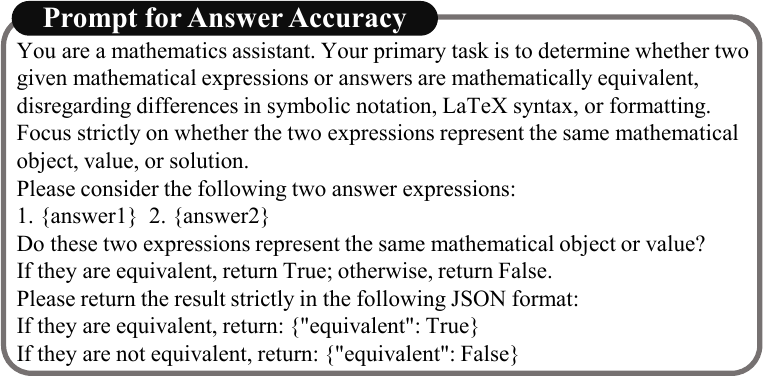}
  \caption{Prompt used for answer accuracy.}
  \label{fig:prompt_answer_acc_app}
\end{figure}

\paragraph{Answer accuracy}
The proportion of answerable questions for which the model produces the correct final answer.
We use the more capable Qwen3-32B model to perform the evaluation, with the specific judgment prompt shown in Figure~\ref{fig:prompt_answer_acc_app}.

\begin{figure*}[!t]
  \centering
  \includegraphics[width=\textwidth]{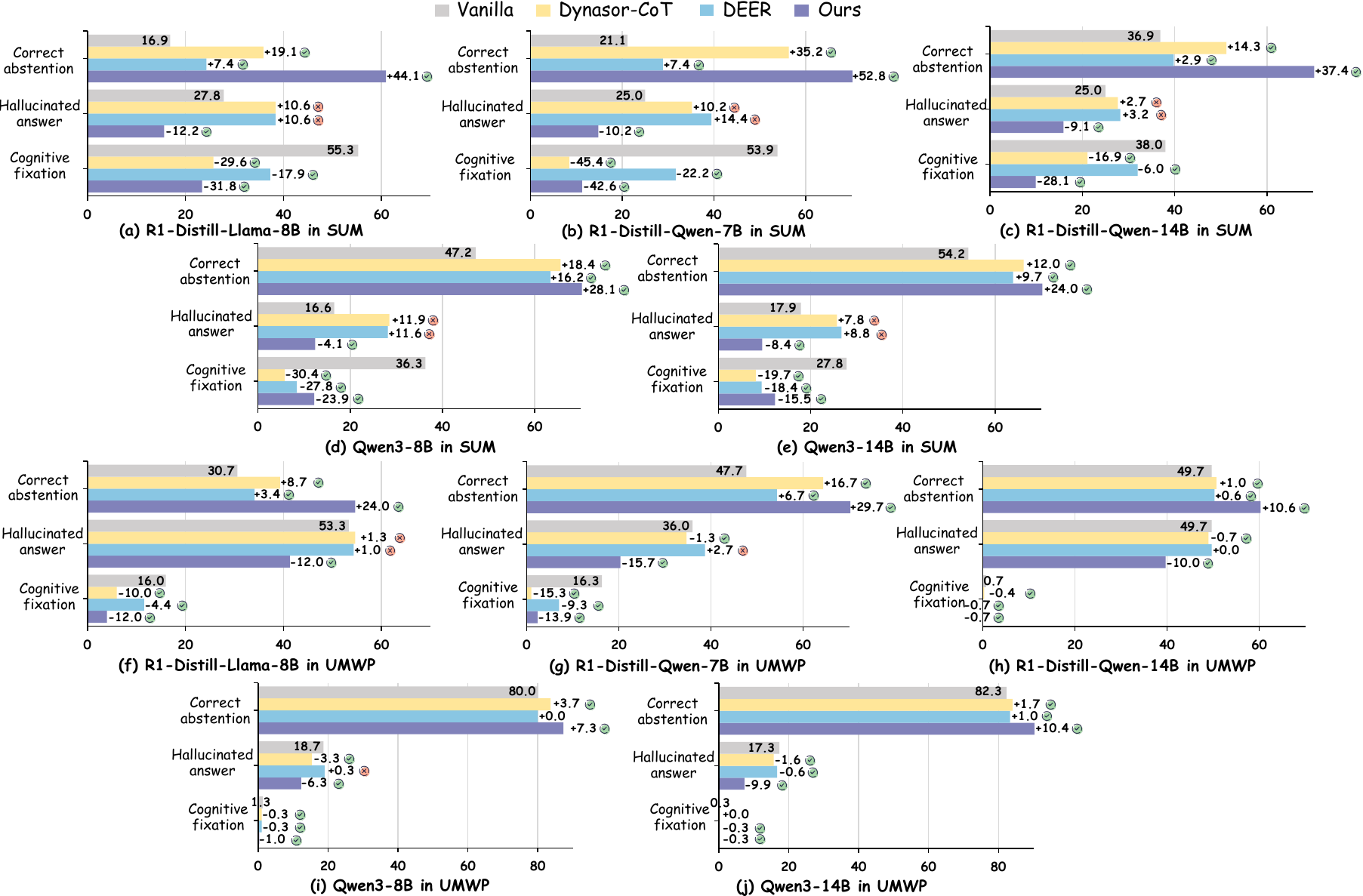}
  \caption{Comparison of response type distributions on unanswerable questions in the SUM and UMWP datasets. Proportions are reported across methods, with numbers indicating absolute changes from the vanilla model.}
  \label{fig:response_distribution_app}
\end{figure*}

\section{Implementation Details about Experiments}
For the baselines, we select two methods base on output consistency (Dynasor-CoT)~\cite{Dynasor-CoT} and confidence (DEER)~\cite{Deer}. 
All the baselines and our method are intervention-based methods during the reasoning process.
We use the keyword ``wait'' as a universal stopping token to trigger the intervention across all methods.
The prompt used to elicit the intermediate answer at the stopping point is: ``\texttt{\textbackslash n **Final Answer**\textbackslash n\textbackslash boxed\{}''. 
For Dynasor-CoT, we prompt intermediate answers and terminate the reasoning process early when the same answer appears three times consecutively.
For DEER, we prompt intermediate answers and terminate the reasoning process early when the confidence of the answer exceeds a predefined threshold (set to 0.95 following the original paper).

Our method consists of two main components.
For Cognitive Monitoring, we train a lightweight linear probe on 2,000 pairs of answerable and unanswerable questions sampled from the SUM dataset~\cite{SUM}, and validate it on an additional 200 pairs.
For the reasoning process of each question, we sample 1,000 token-level activations as input to the probe.
The linear probe is trained for 75 epochs with a batch size of 16,384 and a learning rate of 3e-5 ~\cite{IIT}.
We evaluate the classification performance of the probe using the full reasoning process on both the SUM and UMWP test sets.
The results are shown in Table~\ref{tab:prob_cls_app}, where the F1 score is computed with a threshold of 0.5.
The results demonstrate that the probe achieves good classification performance and generalizes well across datasets.
In our method, the probe is used to predict the answerability of a question based on the generated reasoning content during the inference process. A question is predicted as unanswerable when the probe’s output exceeds a threshold $t$. We set $t$ = 0.6 for the SUM dataset and $t$ = 0.5 for the UMWP dataset.

\begin{table}[!t]
\centering
\resizebox{\columnwidth}{!}{
\begin{tabular}{l|c|cc|cc}
\toprule
\multirow{3}{*}{LRMs} & \multirow{3}{*}{\makecell{Best\\Layer}} & \multicolumn{2}{c|}{SUM} & \multicolumn{2}{c}{UMWP} \\
\cmidrule(lr){3-4}\cmidrule(lr){5-6}
& & AUROC & F1 & AUROC & F1 \\
\midrule
R1-Distill-Llama-8B    & 22 & 0.879 & 0.816 & 0.871 & 0.809  \\
\midrule
R1-Distill-Qwen-7B     & 17 & 0.912 & 0.840 & 0.926 & 0.867  \\
\midrule
R1-Distill-Qwen-14B    & 30 & 0.908 & 0.828 & 0.930 & 0.879  \\
\midrule
Qwen3-8B               & 24 & 0.925 & 0.857 & 0.967 & 0.919  \\
\midrule
Qwen3-14B              & 26 & 0.936 & 0.874 & 0.970 & 0.918  \\
\bottomrule
\end{tabular}
}
\caption{Classification performance of the probe across different LRMs on the SUM and UMWP datasets.}
\label{tab:prob_cls_app}
\end{table}

\section{Full Results for Further Discussions}
\paragraph{Impact on Response Type Distribution.}
We analyze the changes in the proportions of different response types for different methods across all LRMs and datasets, as shown in Figure~\ref{fig:response_distribution_app}. The results are consistent with those presented in Section 5.3 of the main text. 
Our method consistently reduces the hallucinated answers and cognitive fixation outputs. This reduction directly contributes to a substantial increase in the rate of correct abstentions.
While both Dynasor-CoT and DEER employ early-exit strategies to mitigate cognitive fixation, we observe that such baselines often lead to a higher proportion of hallucinated answers. 
Early exits without appropriate guidance may cause the model to make up assumptions or imagined scenarios for giving a definite answer, rather than acknowledging uncertainty. This highlights the importance of combining monitoring with guided intervention to steer LRMs toward proper behavior.

\paragraph{Intervention Effects.}
We assess how our intervention influences the LRM's confidence in abstention and its actual abstention behavior, in order to evaluate whether our method helps bridge the gap between cognitive awareness and abstention behavior.
Specifically, at the intervention point identified by our method, we prompt the LRM to generate intermediate outputs both before and after the intervention. We then measure two key indicators: the confidence when producing ``I don't know'' responses, and the proportion of questions for which the model outputs ``I don't know''. 
The results across all LRMs and datasets are shown in Table~\ref{tab:further_dis_inter_effect_app}.
The results are consistent with those presented in Section 5.3 of the main text.
Our method consistently enhances the confidence in generating abstention responses. 
In addition, the abstention rate also shows corresponding improvements.

\begin{table}[!t]
\centering
\resizebox{\columnwidth}{!}{
\begin{tabular}{l|ll|ll}
\toprule
\multirow{3}{*}{Method} & \multicolumn{2}{c|}{SUM} & \multicolumn{2}{c}{UMWP} \\
\cmidrule(lr){2-3}\cmidrule(lr){4-5}
& Abst. conf. & Abst. rate & Abst. conf. & Abst. rate \\
\midrule
\rowcolor{gray!15} \multicolumn{5}{l}{R1-Distill-Llama-8B} \\
Pre-Interv.     & 79.7 & 30.1\% & 84.1 &41.5\%   \\
Post-Interv.    & 87.3 ($\uparrow$\,9.4\%) & 78.1\% ($\times$\,2.6) & 90.0 ($\uparrow$\,7.0\%) & 81.4\% ($\times$\,1.9)  \\
\midrule
\rowcolor{gray!15} \multicolumn{5}{l}{R1-Distill-Qwen-7B} \\
Pre-Interv.     & 77.1 & 24.5\% & 87.1 &50.8\%   \\
Post-Interv.    & 86.8 ($\uparrow$\,12.6\%) & 80.6\% ($\times$\,3.3) & 92.1 ($\uparrow$\,5.7\%) & 71.5\% ($\times$\,1.4)  \\
\midrule
\rowcolor{gray!15} \multicolumn{5}{l}{R1-Distill-Qwen-14B} \\
Pre-Interv.     & 87.9 & 47.6\% & 89.1 &65.6\%   \\
Post-Interv.    & 91.9 ($\uparrow$\,4.6\%) & 77.0\% ($\times$\,1.6) & 94.9 ($\uparrow$\,6.5\%) & 84.6\% ($\times$\,1.3)  \\
\midrule
\rowcolor{gray!15} \multicolumn{5}{l}{Qwen3-8B} \\
Pre-Interv.     & 90.9 & 48.3\% & 90.6 &75.9\%   \\
Post-Interv.    & 98.9 ($\uparrow$\,8.7\%) & 74.9\% ($\times$\,1.5) & 98.1 ($\uparrow$\,8.2\%) & 93.1\%  ($\times$\,1.2) \\
\midrule
\rowcolor{gray!15} \multicolumn{5}{l}{Qwen3-14B} \\
Pre-Interv.     & 91.2 & 57.8\% & 94.9 &75.6\%   \\
Post-Interv.    & 98.2 ($\uparrow$\,7.7\%) & 94.7\% ($\times$\,1.6) & 98.9 ($\uparrow$\,4.1\%) & 97.3\%  ($\times$\,1.3) \\
\bottomrule
\end{tabular}
}
\caption{Results of intervention effects. ``Abst. conf.'' denotes the average abstention confidence when getting the answer ``I don't know''. ``Interv.'' is the inference-time intervention.}
\label{tab:further_dis_inter_effect_app}
\end{table}

\paragraph{Further Analysis of Cognitive Monitoring.}
We analyze the cognitive monitoring component of our method and compare our default monitoring strategy based on latent representations with alternative strategies relying on behavioral signals.
The behavioral signal approach monitors the LRMs' intermediate outputs at the end of the paragraph generation phase (e.g., when it reaches a ``wait'' token), and uses these outputs to determine whether to trigger an intervention. 
We investigate three variants: 
The ``direct behavior'' strategy checks whether the model's intermediate output is ``I don't know'' and triggers an intervention immediately if so.
The ``consistency'' strategy triggers intervention only if the model produces ``I don't know'' in three consecutive intermediate outputs (inspired by Dynasor-CoT). 
The ``confidence score'' strategy triggers intervention when the model outputs ``I don't know'' with a confidence score exceeding a predefined threshold of 0.95 (inspired by DEER). 
We conduct experiments on two datasets and five different LRMs, and the results are shown in Table~\ref{tab:further_CM_app}. 
The results are consistent with those presented in Section 5.3 of the main text. 
All monitoring strategies contribute to the improvement in abstention behavior, showing that cognitive monitoring is generally effective.
Among them, the strategy based on latent representation signals achieves the best and most consistent performance across different models and datasets. The direct behavior method is simple and works well, but it can be too aggressive and may hurt performance on answerable questions.

\begin{table*}[t]
\centering
\resizebox{\textwidth}{!}{
\begin{tabular}{l|ccc|c|ccc|c}
\toprule
\multirow[b]{3}{*}{Monitoring Signal} & \multicolumn{4}{c|}{SUM} & \multicolumn{4}{c}{UMWP} \\
\cmidrule(lr){2-5}\cmidrule(lr){6-9}
& \multicolumn{3}{c|}{Unanswerable} & \multicolumn{1}{c|}{Ans.} & \multicolumn{3}{c|}{Unanswerable} & \multicolumn{1}{c}{Ans.} \\
\cmidrule(lr){2-4}\cmidrule(lr){5-5} \cmidrule(lr){6-8}\cmidrule(lr){9-9}
& \makecell{Correct\\abstention} $\uparrow$ & \makecell{Hallucinated\\answer} $\downarrow$ & \makecell{Cognitive\\fixation} $\downarrow$ & Acc $\uparrow$ & \makecell{Correct\\abstention} $\uparrow$ & \makecell{Hallucinated\\answer} $\downarrow$ & \makecell{Cognitive\\fixation} $\downarrow$ & Acc $\uparrow$  \\
\midrule
\rowcolor{gray!15} \multicolumn{9}{l}{R1-Distill-Llama-8B} \\
Vanilla               & 16.9 & 27.8 & 55.3 & 61.9 & 30.7 & 53.3 & 16.0 & 77.7  \\
Latent Representation & \textbf{60.9} &\textbf{15.7} & \textbf{23.4} & 60.9 & \textbf{54.7} & \textbf{41.3} & 4.0 & 77.3   \\
Direct Behavior       & 53.1 & 20.4 & 26.5 & 58.1  & 51.0 & 47.0 & \textbf{2.0} & 75.7  \\
Consistency           & 47.9 & 23.6 & 28.5 & 59.8  & 45.3 & 51.0 & 3.7 & \textbf{78.3}  \\
Confidence Score      & 37.0 & 24.7 & 38.4 & \textbf{61.3} & 43.3 & 51.0 & 5.7 & \textbf{78.3}   \\
\midrule
\rowcolor{gray!15} \multicolumn{9}{l}{R1-Distill-Qwen-7B} \\
Vanilla               & 21.1 & 25.0 & 53.9 & 69.7 & 47.7 & 36 & 16.3 & 90.3  \\
Latent Representation & \textbf{73.9} & \textbf{14.8} & \textbf{11.3} & 67.3 & \textbf{77.3} & \textbf{20.3} & \textbf{2.3} & 90.0  \\
Direct Behavior       & 41.6 & 22.6 & 35.9 & 66.8  & 61.3 & 32.0 & 6.7 & 90.0  \\
Consistency           & 35.7 & 23.1 & 41.2 & 69.3  & 57.3 & 34.7 & 8.0 & \textbf{91.0}  \\
Confidence Score      & 31.2 & 23.9 & 44.9 & \textbf{69.4} & 54.3 & 34.3 & 11.3 & 90.3   \\
\midrule
\rowcolor{gray!15} \multicolumn{9}{l}{R1-Distill-Qwen-14B} \\
Vanilla               & 36.9 & 25.0 & 38.0 & 70.4 & 49.7 & 49.7 & 0.7 & 90.0  \\
Latent Representation & \textbf{74.3} & \textbf{15.9} & \textbf{9.9} & 67.9 & \textbf{60.3} & \textbf{39.7} & \textbf{0.0} & 89.7   \\
Direct Behavior       & 67.3 & 19.0 & 13.7 & 64.4 & 53.7 & 46.0 & 0.3 & 90.0   \\
Consistency           & 62.3 & 21.1 & 16.6 & 65.9 & 50.0 & 49.7 & 0.3 & 90.0   \\
Confidence Score      & 53.9 & 22.2 & 23.9 & \textbf{69.4} & 52.0 & 47.7 & 0.3 &  \textbf{91.0} \\
\midrule
\rowcolor{gray!15} \multicolumn{9}{l}{Qwen3-8B} \\
Vanilla               & 47.2 & 16.6 & 36.3 & 60.9  & 80.0 & 18.7 & 1.3 & 94.3  \\
Latent Representation & \textbf{75.3} & 12.4 & \textbf{12.3} & \textbf{61.6}  & 87.3 & 12.3 & \textbf{0.3} & \textbf{93.7} \\
Direct Behavior       & 67.3 & 10.9 & 21.8 & 60.2 & \textbf{91.3} & \textbf{8.3} & \textbf{0.3} & 90.0   \\
Consistency           & 61.6 & 13.0 & 25.4 & 61.3 & 87.3 & 12.3 & \textbf{0.3} & 92.7 \\
Confidence Score      & 64.3 & \textbf{10.6} & 25.2 & 61.3 & 90.3 & 9.3 & \textbf{0.3} & 91.3  \\
\midrule
\rowcolor{gray!15} \multicolumn{9}{l}{Qwen3-14B} \\
Vanilla               & 54.2 & 17.9 & 27.8 & 66.6 & 82.3 & 17.3 & 0.3 & 94.3  \\
Latent Representation & \textbf{78.2} & \textbf{9.5} & \textbf{12.3} & 65.0 & \textbf{92.7} & \textbf{7.3} & \textbf{0.0} & 92.5   \\
Direct Behavior       & 74.3 & \textbf{9.5} & 16.2 & 64.0 & 92.0 & 8.0 & \textbf{0.0} & 90.7   \\
Consistency           & 67.9 & 12.3 & 19.7 & \textbf{66.2} & 87.3 & 12.7 & \textbf{0.0} & \textbf{93.0}   \\
Confidence Score      & 71.5 & 9.9 & 18.7 & 65.1 & 92.0 & 8.0 & \textbf{0.0} & 91.6   \\
\bottomrule
\end{tabular}
}
\caption{Comparison of different cognitive monitoring strategies across different LRMs on SUM and UMWP datasets.}
\label{tab:further_CM_app}
\end{table*}

\paragraph{Ablation Study.}
We evaluate two aspects of inference-time intervention: the instructional guidance prompt and the early exit strategy. 
By removing one component at a time, we analyze how each affects abstention behavior and answer quality. 
The results, shown in Table~\ref{tab:further_dis_ablation_app}, are consistent with those presented in Section 5.3 of the main text.
For correct abstention, the impact of instructional guidance is greater than that of early exit. The early exit strategy helps reduce the number of cognitive fixation cases.
However, without instructional guidance, the proportion of hallucinated answers increases. This again shows that without proper guidance, the model tends to make up conditions and generate unsupported answers. Instructional guidance also has a slight impact on the performance of answerable questions.

\begin{table*}[t]
\centering
\resizebox{\textwidth}{!}{
\begin{tabular}{l|lll|c|lll|c}
\toprule
\multirow[b]{3}{*}{Variant} & \multicolumn{4}{c|}{SUM} & \multicolumn{4}{c}{UMWP} \\
\cmidrule(lr){2-5}\cmidrule(lr){6-9}
& \multicolumn{3}{c|}{Unanswerable} & \multicolumn{1}{c|}{Ans.} & \multicolumn{3}{c|}{Unanswerable} & \multicolumn{1}{c}{Ans.} \\
\cmidrule(lr){2-4}\cmidrule(lr){5-5} \cmidrule(lr){6-8}\cmidrule(lr){9-9}
& \makecell{Correct\\abstention} $\uparrow$ & \makecell{Hallucinated\\answer} $\downarrow$ & \makecell{Cognitive\\fixation} $\downarrow$ & Acc $\uparrow$ & \makecell{Correct\\abstention} $\uparrow$ & \makecell{Hallucinated\\answer} $\downarrow$ & \makecell{Cognitive\\fixation} $\downarrow$ & Acc $\uparrow$  \\
\midrule
\rowcolor{gray!15} \multicolumn{9}{l}{R1-Distill-Llama-8B} \\
Vanilla           & 16.9 & \ \ 27.8 & \ \ 55.3 & \ \ 61.9  & 30.7 & 53.3 & 16.0 & 77.7 \\
Ours              & 60.9 & \ \ 15.7 & \ \ 23.4 & \ \ 60.9  & 54.7 & 41.3 & 4.0  & 77.3 \\
w/o Early Exit      & 43.3 ($\uparrow$ \textbf{26.4}) & \ \ 18.3 ($\downarrow$ \textbf{9.5}) & \ \ 38.4 ($\downarrow$ \textbf{16.9}) & \ \ 61.9  & 45.7 ($\uparrow$ \textbf{15.0}) & 48.7 ($\downarrow$ \textbf{4.6}) & 5.7 ($\downarrow$ 10.3) & 77.0 \\
w/o Instr. Guidance & 26.1 ($\uparrow 9.2$)  & \ \ 33.4 ($\uparrow$ 5.6)   & \ \ 40.5 ($\downarrow$ 14.8) & \ \ \textbf{62.3}  & 41.0 ($\uparrow$ 10.3) & 55.3 ($\uparrow$ 2.0) & 3.7 ($\downarrow$ \textbf{12.3}) & \textbf{78.0}  \\
\midrule
\rowcolor{gray!15} \multicolumn{9}{l}{R1-Distill-Qwen-7B} \\
Vanilla           & 21.1 & \ \ 25.0 & \ \ 53.9 & \ \ 69.7 & 47.7 & 36.0 & 16.3 & 90.3   \\
Ours              & 73.9 & \ \ 14.8 & \ \ 11.3 & \ \ 67.3 & 77.3 & 20.3 & 2.3  & 90.0   \\
w/o Early Exit      & 49.7 ($\uparrow$ \textbf{28.6}) & \ \ 16.2 ($\downarrow$ \textbf{8.8}) & \ \ 34.2 ($\downarrow$ 19.7) & \ \ 69.0 & 67.0 ($\uparrow$ \textbf{19.3}) & 26.0 ($\downarrow$ \textbf{10.0}) & 7.0 ($\downarrow$ 9.3) & 90.3   \\
w/o Instr. Guidance & 43.5 ($\uparrow$ 22.4) & \ \ 35.2 ($\uparrow$ 10.2) & \ \ 21.3 ($\downarrow$ \textbf{32.6}) & \ \ \textbf{70.0}  & 58.0 ($\uparrow$ 10.3) & 40.7 ($\uparrow$ 4.7) & 1.3 ($\downarrow$ \textbf{15.0}) & \textbf{91.0}  \\
\midrule
\rowcolor{gray!15} \multicolumn{9}{l}{R1-Distill-Qwen-14B} \\
Vanilla           & 36.9 & \ \ 25.0 & \ \ 38.0 & \ \ 70.4 & 49.7 & 49.7 & 0.7 & 90.0  \\
Ours              & 74.3 & \ \ 15.9 & \ \ 9.9 & \ \ 67.9  & 60.3 & 39.7 & 0.0 & 89.7  \\
w/o Early Exit      & 63.4 ($\uparrow$ \textbf{26.5}) & \ \ 20.1 ($\downarrow$ \textbf{4.9}) & \ \ 16.6 ($\downarrow$ \textbf{21.4}) & \ \ 69.4 & 55.0 ($\uparrow$ \textbf{5.3})  & 44.7 ($\downarrow$ \textbf{5.0}) & 0.3 ($\downarrow$ 0.4) & \textbf{90.0}  \\
w/o Instr. Guidance & 46.8 ($\uparrow$ 9.9) & \ \ 35.2 ($\uparrow$ 10.2) & \ \ 17.9 ($\downarrow$ 20.1) & \ \ \textbf{70.7} & 51.0 ($\uparrow$ 1.3) & 49.0 ($\downarrow$ 0.7) & 0.0 ($\downarrow$ \textbf{0.7}) & \textbf{90.0}  \\
\midrule
\rowcolor{gray!15} \multicolumn{9}{l}{Qwen3-8B} \\
Vanilla           & 47.2 & \ \ 16.6 & \ \ 36.3 & \ \ 60.9  & 80.0 & 18.7 & 1.3 & 94.3  \\
Ours              & 75.3 & \ \ 12.4 & \ \ 12.3 & \ \ 61.6  & 87.3 & 12.3 & 0.3 & 93.7 \\
w/o Early Exit      & 73.6 ($\uparrow$ \textbf{26.4}) & \ \ 13.7 ($\downarrow$ \textbf{2.9}) & \ \ 12.7 ($\downarrow$ 23.6) & \ \ \textbf{62.7} & 86.3 ($\uparrow$ \textbf{6.3}) & 13.3 ($\downarrow$ \textbf{5.4}) & 0.3 ($\downarrow$ \textbf{1.0}) & 94.0  \\
w/o Instr. Guidance & 59.2 ($\uparrow$ 12.0) & \ \ 30.3  ($\uparrow$ 13.7) & \ \ 10.6 ($\downarrow$ \textbf{25.7})  & \ \ \textbf{62.7} & 80.5 ($\uparrow$ 0.5) & 19.2 ($\uparrow$ 0.5) & 0.3 ($\downarrow$ \textbf{1.0}) & \textbf{94.3}  \\
\midrule
\rowcolor{gray!15} \multicolumn{9}{l}{Qwen3-14B} \\
Vanilla           & 54.2 & \ \ 17.9 & \ \ 27.8 & \ \ 66.6  & 82.3 & 17.3 & 0.3 & 94.3 \\
Ours              & 78.2 & \ \ 9.5 & \ \ 12.3 & \ \ 65.0  & 92.7 & 7.3 & 0.0 & 92.5  \\
w/o Early Exit      & 73.6 ($\uparrow$ \textbf{19.4}) & \ \ 11.9 ($\downarrow$ \textbf{6.0}) & \ \ 14.4 ($\downarrow$ 13.4) & \ \ 65.0  & 92.0 ($\uparrow$ \textbf{9.7}) & 8.0 ($\downarrow$ \textbf{9.3}) & 0.0 ($\downarrow$ \textbf{0.3}) & 92.7  \\
w/o Instr. Guidance & 63.0 ($\uparrow$ 8.8) & \ \ 23.9  ($\uparrow$ 6.0) & \ \ 13.0 ($\downarrow$ \textbf{14.8})  & \ \ \textbf{66.9} & 85.0 ($\uparrow$ 2.7)  & 15.0 ($\downarrow$ 2.3) & 0.0 ($\downarrow$ \textbf{0.3}) & \textbf{94.0} \\
\bottomrule
\end{tabular}
}
\caption{Ablation results of intervention components across different LRMs on SUM and UMWP datasets. We report the effect of removing either Instructional Guidance or Early Exit component on three types of responses for unanswerable questions, as well as the accuracy on answerable questions.}
\label{tab:further_dis_ablation_app}
\end{table*}

\end{document}